\documentclass[11pt]{article}

\usepackage[final]{acl}

\usepackage{times}
\usepackage{latexsym}

\usepackage[T1]{fontenc}

\usepackage[utf8]{inputenc}

\usepackage{microtype}

\usepackage{inconsolata}

\usepackage{graphicx}

\usepackage{booktabs}
\usepackage{amsmath}
\usepackage{listings}
\usepackage{makecell}
\usepackage{subcaption}
\newcommand{\methodname}{\textsc{CLARE}}

\newcommand{\Appendix}{in Appendix }

\usepackage{blkarray}
\usepackage{booktabs}
\usepackage{xcolor}
\usepackage{hyperref}

\title{LVLM-Aware Multimodal Retrieval for RAG-Based Medical Diagnosis with General-Purpose Models}

\author{
 \textbf{Nir Mazor\textsuperscript{1}},
 \textbf{Tom Hope\textsuperscript{1,2}}
\\
\\
 \textsuperscript{1}School of Computer Science and Engineering, The Hebrew University of Jerusalem \\
 \textsuperscript{2}The Allen Institute for AI (AI2)
\\
\raisebox{-0.1\height}{\includegraphics[height=10pt]{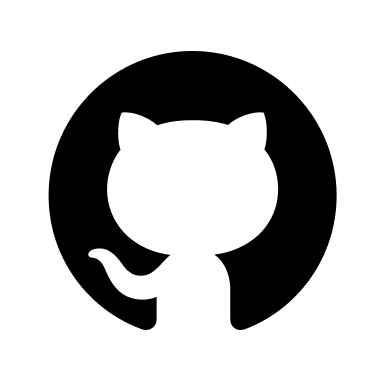}}~\href{https://github.com/Nirmaz/CLARE}{Code and Models}
\\
}

\begin{document}
\maketitle
\begin{abstract}

Retrieving visual and textual information from medical literature and hospital records can enhance diagnostic accuracy for clinical image interpretation. However, multimodal retrieval-augmented diagnosis is highly challenging. We explore a lightweight mechanism for enhancing diagnostic performance of retrieval-augmented LVLMs. We train a lightweight LVLM-aware multimodal retriever, such that the retriever learns to return images and texts that guide the LVLM toward correct predictions. In our low-resource setting, we perform only lightweight fine-tuning with small amounts of data, and use only general-purpose backbone models, achieving competitive results in clinical classification and VQA tasks compared to medically pre-trained models with extensive training. In a novel analysis, we highlight a previously unexplored class of errors that we term inconsistent retrieval predictions: cases where different top-retrieved images yield different predictions for the same target. We find that these cases are challenging for all models, even for non-retrieval models, and that our retrieval optimization mechanism significantly improves these cases over standard RAG. However, our analysis also sheds light on gaps in the ability of LVLMs to utilize retrieved information for clinical predictions.

\end{abstract}

\section{Introduction}
\label{sec:introduction}

\begin{figure}[t!]
    \includegraphics[width=\columnwidth]{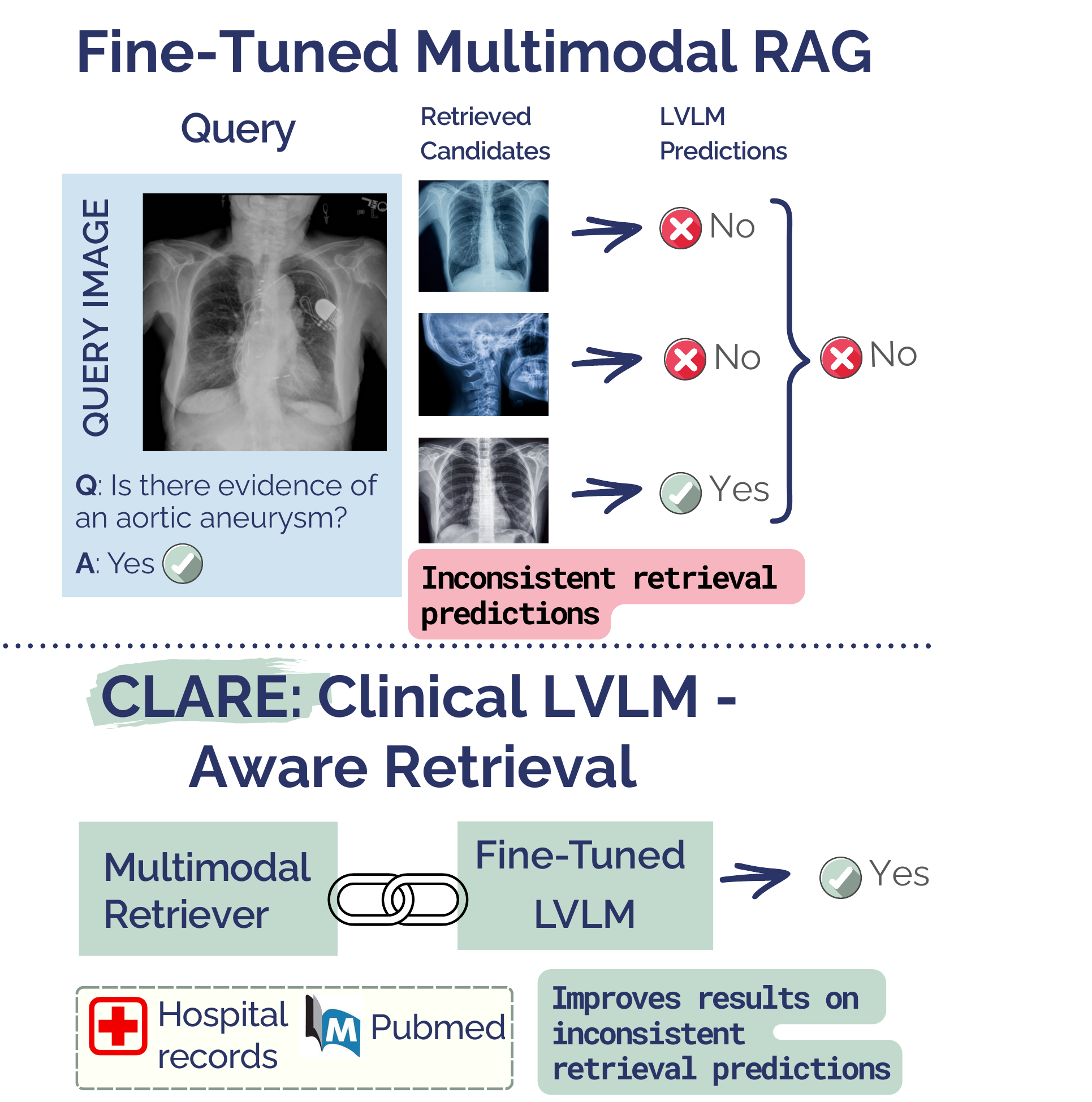}
    \caption{We optimize a multimodal retriever and a Large Vision-Language Model 
    for medical tasks. We achieve competitive results without resource-intensive medical pre-training and significantly improve performance on challenging cases where different retrieved images lead to inconsistent retrieval prediction.}
    \label{fig:motivation}
\end{figure}
Inferring diagnoses from medical imagery is a fundamental part of clinical decision-making. Large Vision Language Models (LVLMs) have been widely explored for medical diagnosis \cite{thawakar2024xraygpt,wu2023towards,zhang2023pmc,moor2023med,li2023llava}. 
To improve the performance of LVLMs in the medical field, retrieval augmentation (RAG) has been adopted and has shown promising results, providing more accurate and also explainable methods \cite{he2024meddr,xia2024rule,xia2024mmed,wu2025mkgf}.

\begin{figure}[t]
\centering
\includegraphics[height=6cm]{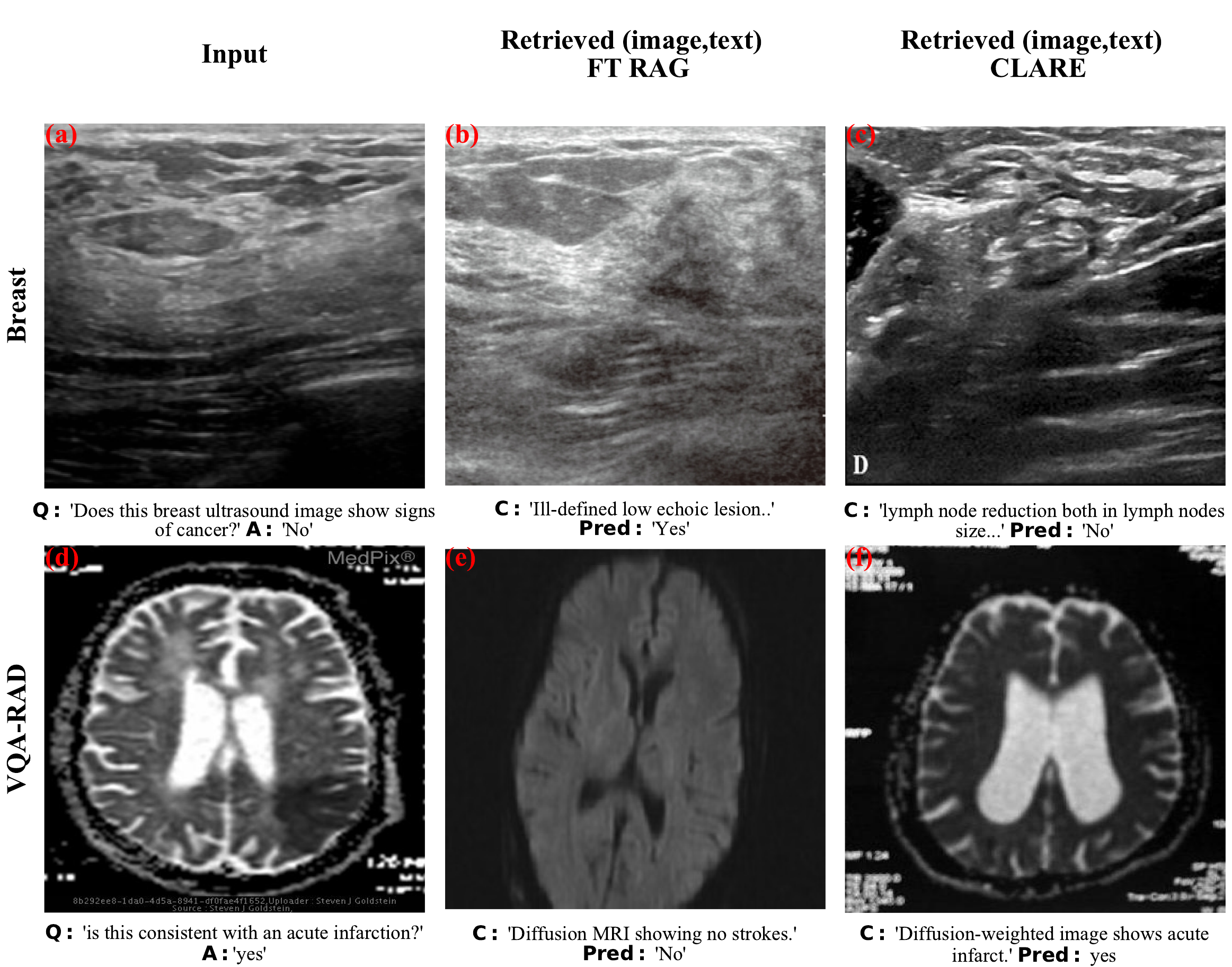}
\caption{{LVLM-aware multimodal retrieval finetuning} impacts inconsistent retrieval predictions. After {LVLM-aware multimodal retrieval finetuning}, retrieved images show greater alignment with query image labels. In Breast, a cancer-free ultrasound query (a) initially retrieved a lesion image (b), however, after finetuning, the retrieved image is less directly related to the wrong label (c). In VQA-RAD, retrieval shifted from an unrelated medical condition (e) to an image depicting the condition of the query image.}
\label{fig:inconsistent_visual}
\end{figure}

In this work, we explore a lightweight fine-tuning approach using a \textit{general-purpose} LVLM with a \textit{general-purpose} multimodal retriever for medical diagnosis tasks. Unlike standard multimodal RAG, we train an LVLM-aware multimodal retriever to find knowledge---clinical images, captions, and reports from both medical literature and hospital records---that leads to correct LVLM predictions. Poorly optimized retrieval mechanisms can mislead models \cite{yoran2023making, sun2024surf}. As shown in Figure \ref{fig:inconsistent_visual}, a RAG model incorrectly classifies a benign ultrasound image when conditioned on a cancerous retrieved image, whereas our LVLM-aware retrieval optimization leads to the correct predicted diagnosis.



Our method involves sequential multimodal training with a dual-head retriever architecture and a customized retrieval loss and visual question answering (VQA) training recipe. 
Our method achieves competitive results in medical image classification and VQA. Importantly, our focus in this work is not to chase SOTA results but to shed light on several interesting observations. One, that a simple and effective LVLM-aware retrieval optimization mechanism can lead to a substantial boost in results over current multimodal RAG methods and should thus be more widely adopted for medical diagnosis; beyond downstream performance, our analysis also shows improved relevance of retrieval candidates. Second, that this mechanism achieves competitive results by using models with no medical pre-training (for neither the LVLM nor the retriever) and only lightweight fine-tuning.  This resonates with recent findings showing general-purpose LVLMs can rival their medical counterparts \cite{jeong2024limited} and has potentially important implications considering the resource-intensive nature of pre-training processes.

Third, related work in the general domain that tuned both retrieval and generation models required large-scale pre-training followed by few-shot task-specific adaptation \cite{izacard2023atlas, hu2023reveal, lin2023ra}. In contrast, we conduct our  optimization \textit{directly on downstream tasks} with no pretraining and only lightweight fine-tuning (as few as 546 samples). This provides first evidence about the feasibility and utility of lightweight data-efficient generation-aware retrieval optimization directly on downstream tasks as opposed to in the pre-training setting. In addition to their reliance on extensive training, another gap in this line of work is that the various techniques developed in the LLM community for tuning retrievers to align with LLMs \cite{izacard2023atlas, shi2023replug, lin2023ra} have yet to be explored in the multimodal setting with LVLMs.


Fourth and importantly, we conduct a novel analysis that finds that our method helps in particular to address a class of errors we term \textit{inconsistent retrieval predictions}. Inconsistent retrieval predictions are cases in which for a given patient image, each retrieved image leads the model to make different predictions. These cases commonly occur across our experiments and are substantially more difficult for models, both retrieval-augmented  and non-retrieval models. This instability with respect to different retrieval candidates leads to high prediction entropy/uncertainty across classes, degrading not only overall RAG results but also their reliability \cite{lambert2022trustworthy}. As we show, our retrieval mechanism significantly mitigates this issue and achieves large improvement over standard fine-tuned RAG on these cases. {Unlike prior work \cite{yoran2023making, sun2024surf}, which focuses on robustness to noisy or irrelevant context without characterizing the target cases, we study a different phenomenon: instances where multiple seemingly relevant candidates induce conflicting predictions. Some candidates may support correct predictions while others mislead the model. Our analysis shows these target cases (input patient images) are inherently difficult, even for non-retrieval models}. We further discover that for a large proportion of these inconsistent cases, at least one retrieved candidate enables the model to predict the correct answer in an oracle setting in which we provide the model with the correct retrieved candidate. This suggests useful information is being retrieved but is often lost among less helpful candidates. To address this, we explored using the powerful o3 model \cite{o32025openai} as a reranker to identify predictive candidates. We found it did not close the performance gap to the oracle and was comparable or inferior to our method, leaving significant room for future improvement.

\textbf{Our contributions} are three-fold: 
\begin{itemize}

    \item  We perform LVLM-aware retrieval optimization for a multimodal retriever and an LVLM on medical classification and VQA tasks (including text generation VQA). We use only general-purpose backbones without medical pretraining, and show that by retrieving relevant medical evidence and training the retriever to guide the LVLM, we close much of the gap to expensive medically pre-trained models and achieve superior performance compared to general-purpose RAG methods.

    \item Unlike previous general-domain methods that required pre-training, we conduct lightweight data-efficient optimization of a retriever and LVLM directly on downstream tasks. 

    
    \item We identify \textit{inconsistent retrieval predictions}, challenging cases where retrieved images lead to different predictions for the same query. Our method significantly improves performance on these cases over standard RAG.
    
\end{itemize}

\section{Related Work}
\label{sec:related_work}

\textbf{Retrieval-Augmented LLM-Aware Optimization} has been explored primarily in unimodal (text only) NLP with work such as ATLAS \cite{izacard2023atlas}, which conducted large-scale pretraining of both reader and retriever models and evaluated in zero-shot and few-shot scenarios in the general domain. REVEAL \cite{hu2023reveal} extended this to the multimodal setting in the general domain with extensive pre-training; REVEAL used an encoder-decoder architecture with a T5 generator---to our knowledge, since REVEAL no work has explored optimizing multimodal retrievers and generators with causal decoder models and modern VLMs. \citet{shi2023replug} proposed training textual retrievers based on reader performance, influencing subsequent retrieval alignment LLM-aware optimization methods. \citet{lin2023ra} introduced another LLM-aware optimization variant and proposed first training the reader with retrieval augmentation, then training the retriever. 
\citet{siriwardhana2023improving} demonstrated LLM-aware optimization's effectiveness for domain adaptation. Our work differs by being the first to explore generator-aware retrieval optimization directly for downstream tasks, with minimal fine-tuning, as opposed to requiring large-scale pre-training. Prior work, such as ATLAS and REVEAL, follows a two-step approach—first conducting large-scale pre-training, then evaluating in few-shot or zero-shot settings, while we use a lightweight one-step approach instead. Our method includes a new loss and new training methodologies. Our novel analyses are the first to shed light on inconsistent predictions and methods to help address them. 

\textbf{Multimodal Retrieval Augmentation in Medical Applications.} Multimodal retrieval augmentation began with encoder-based architectures \cite{yuan2023ramm}, which integrated retrieved text and images with query modalities. With LVLMs, \citet{he2024meddr} proposed retrieving labeled examples during inference. \citet{xia2024rule} improved LVLM RAG with contrastive learning and context selection strategies. \citet{xia2024mmed} proposed domain-aware retrieval for diverse medical data, while \citet{wu2025mkgf} combined retrieval augmentation with knowledge graphs. All current multimodal medical RAG methods are not LVLM-aware, and the retriever is trained disjointly from the LVLM.

\section{Method}
\label{sec:Method}

\begin{figure*}[t]
    \centering
    \includegraphics[width=\textwidth]{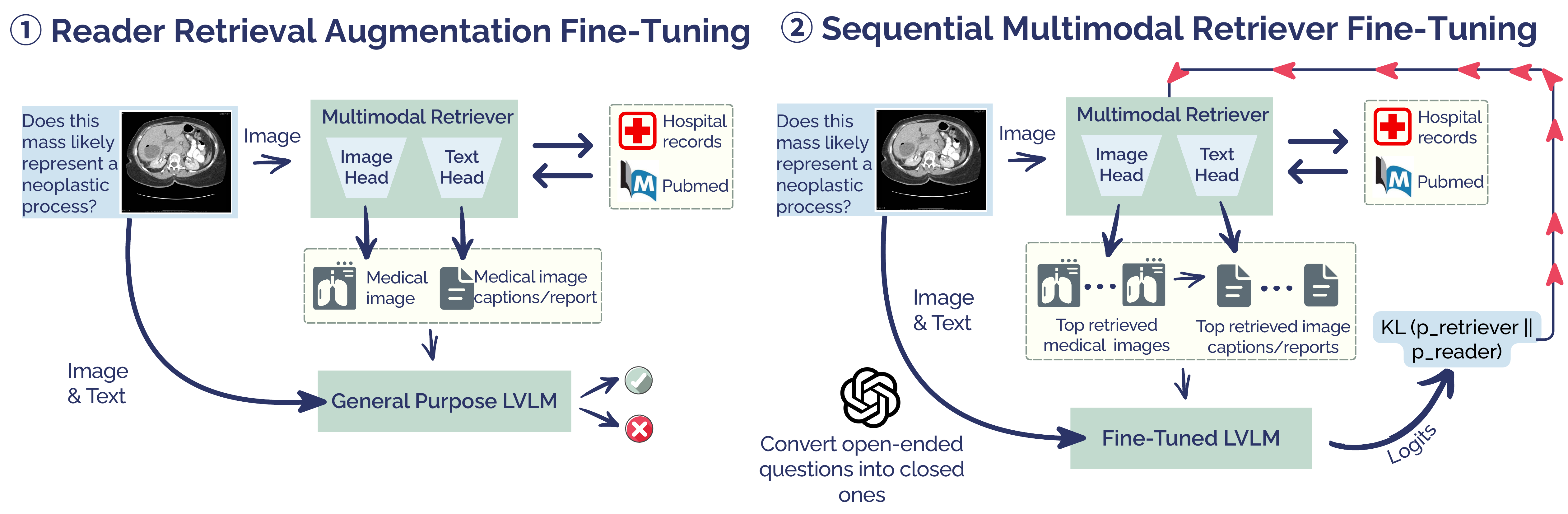}
        \caption{The two-phase \methodname{} training. The LVLM is first trained with a frozen retriever on augmented prompts containing retrieved image–caption/report pairs. With the LVLM frozen, the dual-head multimodal retriever is optimized using a KL divergence computed over a selected subset of the model’s logits. An LLM converts open-ended questions into closed form (during training only), which improves results.
}
   
    \label{fig:method}
\end{figure*}

\textbf{Overview.} {In this section, we present our methodology, termed \methodname{} ({Clinical LVLM-Aware Retrieval}). Our task involves medical image classification and visual question answering. \methodname{} comprises two main components: a multimodal retriever and a reader. Given an input patient image and a diagnostic question, the multimodal retriever identifies relevant medical knowledge—images and their associated captions or hospital reports—that provide predictive information. The reader then analyzes the retrieved candidates along with the question to generate an answer. We optimize the reader to better leverage the retrieved information and the multimodal retriever to enhance its ability to select informative candidates for the reader. {Figure~\ref{fig:method} illustrates our framework.} }


Unlike previous work \cite{he2024meddr, xia2024rule,xia2024mmed} we do not use medical pretraining, and instead we use readily available general LVLMs (Pixtral and Qwen2-vl) and a general retriever, jina-clip \cite{xiao2024jina}. {The selected LVLMs can process multiple images and texts, allowing us to incorporate both the retrieved image and the retrieved text}. We also compare to Med-Flamingo \cite{moor2023med}, a medical LVLM which is also able to process multiple images, unlike other medical LVLMs available at the time of conducting our experiments. We now elaborate on the details.

\subsection{Reader Fine-Tuning}
\label{finetunning}
We first fine-tune the reader with retrieval augmentation. This step aims to achieve two main goals: improving the reader's performance on the dataset and teaching the reader to effectively utilize retrieved (image, text) pairs in context.

For an image-question tuple $(i_d, q_d)$ where $d\in \{1, \ldots, D\}$ from dataset $D$, we retrieve $K$ relevant tuples of images and texts (reports or captions) $(i_k, t_k)$ where $k\in \{1, \ldots, K\}$ using our multimodal retriever module. To simplify notation, we denote $z_k = (i_k, t_k)$ as the $k$-th retrieved image-text pair and $q_d = (i_d, q_d)$ as the query pair for example $d$. These retrieved pairs are prepended to the query pair to create augmented inputs. The augmented input consists of the retrieved context followed by the target question-image pair: $z_k \circ q_d$ for $k\in \{1, \ldots, K\}$. We use supervised fine-tuning on the augmented input, minimizing the negative log-likelihood to predict the answer $a_d$:
\[
\mathcal{L}(\theta) = -\sum_{d=1}^{D}\sum_{k=1}^{K}\log p_{\theta}(a_d \mid z_k \circ q_d)
\]
where $\theta$ represents the parameters of the LVLM, and $\circ$ denotes the concatenation operation. 





\subsection{Sequential Retriever Fine-Tuning}

{Our dual-head multimodal retriever has a text-based head for retrieving relevant (image, text) pairs based on textual similarity to the query, and an image-based head for visual similarity. We adopt a sequential training strategy, first optimizing the text retrieval head followed by the image head. During this process, the reader remains frozen, solely on improving retriever's embedding space. Similarly to \citet{izacard2023atlas} we minimize the KL divergence between the LVLM's posterior distribution over retrieved pairs and the retriever's distribution, however we do so directly on downstream tasks with only lightweight fine-tuning in the data-efficient regime without pre-training, and we customize the loss and the training recipe for open-ended questions leading to empirical gains.}




Let $z_k = (\mathbf{i}_k, \mathbf{t}_k)$ denote the $k$-th retrieved image-text pair and $q = (\mathbf{i}, \mathbf{q})$ denote the query pair. The posterior distribution reflects the model's confidence in predicting the correct answer $\mathbf{a}$ given a retrieved pair $z_k$ and query $q$: $p_k \propto p_{\text{LVLM}}(\mathbf{a} \mid z_k \circ q)$, where $p_{\text{LVLM}}(\mathbf{a} \mid z_k \circ q)$ is the probability assigned by the LVLM to $\mathbf{a}$, and $\circ$ denotes prepending the retrieved chunk to the query.

We sharpen the LVLM's output distribution by restricting it to only the relevant class tokens from the benchmark's possible answers. Specifically, we extract the model's logits, select only the relevant tokens for each class, and apply a softmax to obtain a more discriminative distribution. We denote this class-restricted distribution as $p_{\text{LVLM}_C}$ where $C = \{c_1, c_2, c_3, \ldots\}$ is the set of chosen class tokens. The normalized posterior $p_k$ is formulated as:

\[
p_k = \frac{\exp(\log p_{\text{LVLM}_C}(\mathbf{a} \mid z_k \circ q))}
{\sum_{i=1}^{K} \exp(\log p_{\text{LVLM}_C}(\mathbf{a} \mid z_i \circ q))},
\]

where $K$ is the total number of retrieved pairs.

The retriever's distribution is defined by:

\[
p_{\text{RETR}}(z \mid q) = 
\frac{\exp(s(z, q) / \tau)}
{\sum_{k=1}^{K} \exp(s(z_k, q) / \tau)},
\]

where $s(z, q)$ is the similarity score between the pair and the query, and temperature $\tau$ controls distribution sharpness. For the text retrieval head, the score is computed using the dot product between the index caption/report embeddings and the query image embedding, while for the image retrieval head, we use the index image embeddings. 

For our loss we compute the KL-divergence between retriever and reader  $\text{KL}(p_{\text{LVLM}_C} \| p_{\text{RETR}})$:

$$\sum_{k=1}^{K} p_{\text{LVLM}_C}(z_k) \log \left( \frac{p_{\text{LVLM}_C}(z_k)}{p_{\text{RETR}}(z_k)} \right)$$

{Finally, for visual question answering (VQA), we use the \texttt{o3} model to convert open-ended questions into closed-ended ones (yes/no). See prompt in Appendix~\ref{prompt_design} and analysis of approach contribution \Appendix\ref{model_comp}. We apply this only during training\footnote{Run once per question; the total cost is only a few dollars.}}, without modifying questions at inference time. Reformatted questions will be made publicly available. This reformulation improves results, akin to restricting the LVLM output distribution.



\subsection{Inference}
\label{inference_stage}
Following \citet{shi2023replug} for a given query pair $q = (\mathbf{i}, \mathbf{q})$, we retrieve the top-$K$ relevant chunks $z_k = (\mathbf{i}_k, \mathbf{t}_k)$ using the image as the query. Each chunk is prepended to the question, and the LVLM computes predictions for the augmented prompts in parallel. The final output probability is:

\[
p_{\text{LVLM}}(a \mid q) = \sum_{k=1}^{K} p_{\text{LVLM}}(a \mid z_k \circ q) \cdot p_{\text{R}}(z_k \mid q),
\]

where $p_{\text{R}}(z_k \mid q)=\frac{\exp(s(q, z_k))}{\sum_{j=1}^{K} \exp(s(q, z_j))}$, and $s(q, z_k)$ is the similarity score between the query and the retrieved chunk.

\begin{table*}[t!]
\footnotesize
\begin{subtable}{\textwidth}
\centering
\caption{}
\label{tab:full_results_classification}
\begin{tabular}{@{}lc|cc|cc|cc|cc|cc|cc@{}}
\toprule
Backbone (Size) &  Model & \multicolumn{2}{c}{Breast} & \multicolumn{2}{c}{Derma} & \multicolumn{2}{c}{Retina} & \multicolumn{2}{c}{VinDr-PCXR} & \multicolumn{2}{c}{BRSET} & \multicolumn{2}{c}{Mean} \\
\cmidrule(lr){3-4} \cmidrule(lr){5-6} \cmidrule(lr){7-8} \cmidrule(lr){9-10} \cmidrule(lr){11-12} \cmidrule(lr){13-14}
& & ACC & F1 & ACC & F1 & ACC & F1  & ACC & F1 & ACC & F1 & ACC & F1 \\ 
\midrule
   LLaVA (7B) & MMed-RAG & .85 & .84 & .75 & .30 & .63 & .46 & .55 & .11 & .42 & .30 & .64 & .40 \\ \midrule
Qwen2-vl (7B) &   Reader & .83 & .77 & .68 & .27 & .60 & .44 & .48 & .08 & .34 & .23 & .59 & .36 \\
            &   RAD    & .84 & .79 & .43 & .34 & .55 & .40 & .57 & .09 & .40 & .25 & .56 & .37 \\ 
            &   FT RAG & .85 & .82 & .71 & .42 & .62 & .48 & .55 & .09 & .48 & .27 & .64 & .42 \\ 
            &   \methodname{} + PDist  & .86 & .83 & .73 & .47 & .63 & .49 & .57 & .09 & .47 & .29 & .65 & .43   \\ \cmidrule(lr){2-14}
            &    \methodname{}  & \textbf{.87} & \textbf{.84} & \textbf{.76} & \textbf{.50} & \textbf{.65} & \textbf{.50} & \textbf{.57} & \textbf{.14} & \textbf{.49} & \textbf{.37} & \textbf{.67} & \textbf{.47} \\ \cmidrule(lr){2-14}    
            &    \methodname{}\textsubscript{oracle}  & .87 & .84 & .85 & .64 & .69 & .51 & .64 & .14 & .52 & .41 & .71 & .51 \\ \midrule    
         
Pixtral (12B) &    Reader & .82 & .77 & .75 & .52 & .55 & .44 & .48 & .08 & .44 & .33 & .61 & .43 \\
         &    RAD    & .85 & .80 & .75 & .47 & .56 & .41 & .48 & .09 & .47 & .30 & .62 & .41 \\ 
         &    FT RAG & .88 & .85 & .79 & .60 & .57 & .47 & .49 & .09 & .47 & .33 & .64 & .47 \\ 
         &   \methodname{} + PDist   & .88 & .84 & .80 & .62 & .60 & .47 & .50 & .09 & .45 & .35 & .65 & .47   \\ \cmidrule(lr){2-14}
         &    \methodname{}  & \textbf{.90} & \textbf{.87} & \textbf{.80} & \textbf{.62} & \textbf{.60} & \textbf{.51} & \textbf{.56} & \textbf{.14} & \textbf{.51} & \textbf{.37} & \textbf{.67} & \textbf{.50} \\ \cmidrule(lr){2-14}
         &     \methodname{}\textsubscript{oracle}  & .93 & .90 & .83 & .67 & .63 & .54 & .67 & .15 & .57 & .41 & .73 & .53 \\ 
         
\bottomrule
\end{tabular}
\end{subtable}

\begin{subtable}{\textwidth}
\centering
\caption{}
\label{tab:full_results_vqa}
\begin{tabular}{@{}lc|cc|cc|cc|cc@{}}
\toprule
Backbone (Size) & Model & \multicolumn{2}{c}{VQA-RAD} & \multicolumn{2}{c}{SLAKE} & \multicolumn{2}{c}{PathVQA} & \multicolumn{2}{c}{Mean} \\
\cmidrule(lr){3-4} \cmidrule(lr){5-6} \cmidrule(lr){7-8} \cmidrule(lr){9-10}
& & Closed & Open & Closed & Open & Closed & Open & Closed & Open   \\ \midrule
LLaVA (7B) &     MMed-RAG  &  .74 & .39 & .87 & .81 & .90 & .31 & .84 & .50 \\ 
\midrule
Qwen2-vl (7B) &    Reader  &  .73 & .41 & .84 & .80 & .87 & .25 & .81 & .49 \\
         &     FT RAG     & .76 & .45 & .88 & .81 & .91 & .33 & .85 & .53 \\
         &     \methodname{} + PDist   & .77 & .45 & .89 & .81 & .92 & .32 & .86 & .53 \\
         &    \methodname{}  & \textbf{.79} & \textbf{.48} & \textbf{.90} & \textbf{.84} & \textbf{.93} & \textbf{.38} & \textbf{.87} & \textbf{.57} \\ \cmidrule(lr){2-10}
         &    \methodname{}\textsubscript{oracle}  & .86 & - & .92 & - & .95 & - & .91 & - \\ \midrule
         
Pixtral (12B)  &    Reader & .72 & .38 & .83 & .80 & .87 & .26 & .81 & .48 \\
          &    FT RAG  & .74 & .41 & .88 & .81 & .88 & .31 & .83 & .51  \\
          &    \methodname{} + PDist   & .75 & .41 & .89 & .81 & .89 & .32 & .84 & .51 \\ 
         &     \methodname{}  & \textbf{.76} & \textbf{.45} & \textbf{.90} & \textbf{.84} & \textbf{.90} & \textbf{.36} & \textbf{.85} & \textbf{.55} \\ \cmidrule(lr){2-10}
         &     \methodname{}\textsubscript{oracle}  & .88 & - & .92 & - & .95 & - & .92 & - \\ 
     
\bottomrule
\end{tabular}
\end{subtable}
\caption{Results for classification (a) and VQA (b) with non-medically pretrained LVLM backbones (see Tables \ref{tab:binary_multi_class_table}, \ref{tab:compare_vqa} for comparisons to medical pre-training; MMed-RAG's retrieval component is medically pre-trained). \methodname{} consistently leads to the best results. For VQA, we evaluate closed questions using the exact match metric and open questions using token recall. Using an oracle reranker (for classification and closed VQA; §\ref{sec:Results}) demonstrates that even larger improvements are achievable with \methodname{}'s top-retrieved images in some datasets.   
}
\end{table*}

\begin{table}
\centering
\scriptsize
\setlength{\tabcolsep}{1mm} 

\begin{tabular}{@{}lccccc@{}}
\toprule
                    & MMed-RAG & \makecell{LLaVaMed\\variants} & \makecell{MedDr\\variants} & GSCo & \makecell{MedVInT\\variants}    \\

\midrule
\makecell{Medical Pre-Training\\Size} &  600K    & 600K                         & 255K          & 255K & 177K   \\
\bottomrule
\end{tabular}%
\footnotesize
\caption{Medical pre-training sizes of existing LVLMs.}
\label{tab:medical_pretraining}
\end{table}

\section{{Experiments}}
\label{sec:Expermints}

\begin{table}[t!]
\centering
\footnotesize
\setlength{\tabcolsep}{1mm} 
\begin{tabular}{@{}l|cc|cc@{}}
\toprule
 Model & Breast & Derma & VinDr & BRSET \\ 

 \cmidrule(lr){2-2} \cmidrule(lr){3-3} \cmidrule(lr){4-4} \cmidrule(lr){5-5}  
 MMed-RAG\textsubscript{LLaVA-Med} & .89 & .79 & .11 & .33 \\ \midrule
 MedVInT-TE & .88 & .78 & - & -  \\ \midrule
 MedVInT-TD & .90 & .80  & - & - \\ \midrule
MedDr\textsubscript{InternVL} & .72 & - & .08 & .08   \\ \midrule
 MedDr + RAD\textsubscript{InternVL} & .88 & - & - & -  \\ \midrule
  GSCo\textsubscript{InternVL} & .93 & - & .09 & .33  \\ \midrule
\methodname{} \textsubscript{Qwen2-vl} & .87 & .76 & .14 & .37 \\\midrule
\methodname{}\textsubscript{Pixtral}   & .90 & .80 & .14 & .37  \\
\bottomrule
\end{tabular}
\caption{Comparison to prior reported results of medical pre-trained models, for binary and multiclass classification (Breast and Derma; accuracy), and for multi-label classification (VinDr-PCXR and BRSET; F1). }
\label{tab:binary_multi_class_table}
\end{table}

\begin{table}
\centering
\footnotesize
\setlength{\tabcolsep}{1mm} 
\begin{tabular}{@{}l|cc|cc@{}}
\toprule
Model & \multicolumn{2}{c}{SLAKE} & \multicolumn{2}{c}{PathVQA} \\ 

\cmidrule(lr){2-3} \cmidrule(lr){4-5}  
 &  Closed & Open & Closed & Open    \\ \midrule

MMed-RAG\textsubscript{LLaVA-Med}   &  .89 & .84  & .92 & .39 \\ \midrule
 BiomedGPT-B & .90 & \textcolor{red}{.85} & .88 & \textcolor{red}{.28} \\ \midrule
LLaVA-Med\textsubscript{LLaVA}   &  .83 & .85  & .91 & .38 \\ \midrule
LLaVA-Med\textsubscript{Vicuna}   &  .85 & .83 &  .92 & .39  \\ \midrule
 LLaVA-Med\textsubscript{BioMedCLIP}   & .87 & .87 &  .91 & .39 \\ \midrule
\methodname{}\textsubscript{Qwen2-vl}   & .90 & .84 (\textcolor{red}{.84}) &   .93 & .38 (\textcolor{red}{.37})   \\ \midrule
\methodname{}\textsubscript{Pixtral}   &  .90 & .84 (\textcolor{red}{.82})& .90 & .36 (\textcolor{red}{.35})  \\

\bottomrule
\end{tabular}
\caption{Comparison to medical pre-trained models for visual question answering. We report the token recall metric, reported for LLaVA-Med variants and token F1 reported for BiomedGPT (in \textcolor{red}{red}).}
\label{tab:compare_vqa}
\end{table}




\subsection{Experimental Setup}
\label{setup}





Our datasets include real-world hospital datasets (BRSET \cite{nakayama2024brazilian} and VinDr‑PCXR \cite{pham2022vindr}) alongside a variety of classification and VQA datasets. We focus on a low-resource data-efficient setting (training sets ranging from 546 to 7007 samples in classification, 1790-19,700 in VQA). Medical image annotation is a resource-intensive task, demanding expert annotators whose availability is limited and expensive. Clinical AI also often has poor generalization across heterogeneous medical centers and patient populations, often requiring that models be trained for each deployment \cite{epic2022}.
Such constraints often restrict researchers to datasets comprising only a few thousand samples for a given study. Retrieval augmentation is well-suited for this setting, as it is known to benefit low‑data regimes the most by leveraging external knowledge. 


We consider binary, multi-class, and multi-label classification. These include BreastMNIST (``Breast''; breast ultrasound imaging, binary) \cite{al2020dataset}, DermaMNIST (``Derma''; pigmented skin lesion images, multi-class) \cite{tschandl2018ham10000}, RetinaMNIST (``Retina''; retinal fundus images, multi-class) \cite{liu2022deepdrid}, and two challenging real-world multi-label datasets: VinDr-PCXR (chest X-rays, 15 labels) \cite{pham2022vindr} and BRSET (ophthalmology, 14 labels) \cite{nakayama2024brazilian}. For VQA, we use widely adopted benchmarks: VQA‑RAD \cite{lau2018dataset}, SLAKE‑English \cite{liu2021slake}, and PathVQA \cite{he2020pathvqa}. Full details are provided \Appendix \ref{eval_datasets}.

{Retrieval augmentation is supported by an external index constructed from PubMed and medical records: PMC‑OA \cite{lin2023pmc}, MIMIC‑CXR \cite{johnson2019mimic}, and ROCO \cite{ruckert2024rocov2}. Full descriptions are available \Appendix \ref{Index_datasets}. Our experiments leverage two backbone models—Pixtral (12B) \cite{agrawal2024pixtral} and Qwen2‑vl (7B parameters) \cite{wang2024qwen2}—with jina‑clip‑v1 \cite{xiao2024jina} serving as the general-purpose retriever’s visual head, embedding both the texts and images of the retrieval index. All runs were performed on a single L40s GPU. More details \Appendix \ref{implementation_details}.}


\textbf{Baselines.} {We compare to several RAG baselines. RAD \cite{he2024meddr}, a retrieval-based approach for classification, where the label of the most similar training image is used as context; we use RAD with both Qwen2-VL and Pixtral. MMed-RAG \cite{xia2024mmed}, a recent state-of-the-art multimodal RAG framework in which the retriever is optimized independently from the LVLM and is not aware of the LVLM during training. We evaluated MMed-RAG using backbones LLaVA-Med and LLaVA, both paired with a medically pre-trained retriever (CLIP). We also adopt a standard fine-tuned RAG baseline in which the retriever is fixed and only the reader is fine-tuned, corresponding to the first phase of our approach. Finally, we compare our retriever loss with the base Perplexity Distillation Loss (PDist) \cite{izacard2023atlas}, integrated into \methodname{} in place of our customized loss. Implementation details are in Appendix~\ref{sec:baselines}}. To compare to medically pre‑trained LVLMs, we include competitive baselines: BiomedGPT \cite{luo2023biomedgpt}; three LLaVA-Med \cite{li2023llava} variants; two LVLM variants based on PMC CLIP (MedVInT-TE and MedVInT-TD) \cite{zhang2023pmc}; and three InternVL\cite{chen2024internvl} variants: MedDr, MedDr + RAD \cite{he2024meddr}, GSCo \cite{he2024gsco}. Details \Appendix \ref{sec:baselines}.
 
\subsection{Results}
\label{sec:Results}

{Table~\ref{tab:full_results_classification} and Table~\ref{tab:full_results_vqa} show the effectiveness of \methodname{} on five medical classification and three VQA benchmarks. We compare with methods of similar computational cost; all LVLMs were not extensively pre-trained on medical data, whereas the MMed-RAG retriever was. \methodname{} consistently outperforms MMed-RAG, RAD, the fine-tuned RAG baseline, \methodname{} + PDist loss, and the Reader baseline (LVLM without retrieval).}
In addition, we evaluated our model in an oracle scenario. Instead of the final fusion of logits (Section \ref{inference_stage}), we applied an oracle that, given the model responses for different retrieved images, selects the correct answer if it exists.\footnote{For open-ended questions we do not conduct this analysis as there is no simple boundary between correct/incorrect.} Our findings in Tables \ref{tab:full_results_classification} and \ref{tab:full_results_vqa} show that the correct answer exists in a large proportion of cases, yielding superior results. This shows that in these cases, \methodname{} is able to surface in its top retrieved images an image which could lead to a correct prediction, however the information gets lost in the process of prediction fusion with other retrieved images. Simply taking the label with highest confidence or label with highest average confidence, underperforms. This motivates us to explore the o3 multimodal reasoning model to detect this image (Section \ref{rerank_analysis}).

\begin{table*}
\centering
\footnotesize
\setlength{\tabcolsep}{1mm} 
\begin{tabular}{@{}lcc|cc|cc|cc|cc|cc|c|c|c|c@{}}
\toprule
Backbone &  & Model & \multicolumn{2}{c}{Breast} & \multicolumn{2}{c}{Derma} & \multicolumn{2}{c}{Retina} & \multicolumn{2}{c}{VinDr} & \multicolumn{2}{c}{BRSET} & VQA-RAD & SLAKE  & \multicolumn{1}{c}{\textbf{Mean}} & \multicolumn{1}{c}{\textbf{Mean}} \\
\cmidrule(lr){4-5} \cmidrule(lr){6-7} \cmidrule(lr){8-9} \cmidrule(lr){10-11} \cmidrule(lr){12-13} \cmidrule(lr){14-14} \cmidrule(lr){15-15} \cmidrule(lr){16-16} \cmidrule(lr){17-17}  
& & & ACC & F1 & ACC & F1 & ACC & F1  & ACC & F1 & ACC & F1 & \makecell{Closed\\Acc} & \makecell{Closed\\Acc}  & \textbf{ACC} & \textbf{F1} \\
Qwen2-vl                & Inconsistent & Reader  & .40 & .40 & .50 & .25 & .35 & .31 & .41 & .07 & .33 & .21 & .43 & .70 & .40 & .25 \\  
& &                                          RAG & .40 & .40 & .52 & .36 & .40 & .28 & .46 & .07 & .49 & .25 & .33 & .84 & .45 & .27 \\ 
& & \methodname{}                                & \textbf{.80} & \textbf{.80} & \textbf{.62} & \textbf{.46} & \textbf{.52} & \textbf{.29} & \textbf{.48} & \textbf{.16} & \textbf{.50} & \textbf{.36} & \textbf{.47} & \textbf{.86} & \textbf{.58} & \textbf{.41} \\\cmidrule(lr){2-17}
& Consistent & Reader                            & .84 & .77 & .83 & .29 & .63 & .45 & .56 & .09 & .34 & .22 & .77 & .85 & .64 & .36  \\  
                               & &RAG            & .86 & .82 & .90 & .49 & \textbf{.66} & .50 & \textbf{.66} & \textbf{.10} & \textbf{.43} & .23 & \textbf{.82} & .89 & \textbf{.70} & .43  \\ 
                        & &  \methodname{}       & \textbf{.88} & \textbf{.84} & \textbf{.91} & \textbf{.50} & \textbf{.66} & \textbf{.55} & \textbf{.66} & \textbf{.10} & .41 & \textbf{.37} & \textbf{.82} & \textbf{.91} & \textbf{.70} & \textbf{.47} \\  \midrule
                        
Pixtral                 & Inconsistent & Reader  & .60 & .59 & .45 & .33 & .40 & .32 & .38 & .07 & .46 & .28 & .66 & .64 & .46 & .32 \\  
& &                                          RAG & .72 & .71 & .44 & .45 & .35 & .37 & .38 & .08 & .39 & .26 & .72 & .65 & .46 & .37 \\ 
& & \methodname{}                                & \textbf{.84} & \textbf{.84} & \textbf{.50} & \textbf{.50} & \textbf{.42} & \textbf{.46} & \textbf{.49} &\textbf{.13} & \textbf{.45} & \textbf{.35} & \textbf{.77} & \textbf{.68} & \textbf{.54} & \textbf{.46} \\\cmidrule(lr){2-17}
& Consistent & Reader                            & .86 & .80 & .80 & .56 & .56 & .44 & .66 & .08 & .61 & .37 & .74 & .85 & .70 & .45  \\  
                               & &RAG            & \textbf{.91} & \textbf{.87} & .84 & \textbf{.67} & \textbf{.62} & .50 & \textbf{.71} & .08 & \textbf{.64} & .38 & \textbf{.75} & .91 & .74 & .50  \\ 
                          & &  \methodname{}     & \textbf{.91} & \textbf{.87} & \textbf{.85} & \textbf{.67} & \textbf{.62} & \textbf{.53} & \textbf{.71} & \textbf{.10} & \textbf{.64} & \textbf{.42} & \textbf{.75}  & \textbf{.91} & \textbf{.75} & \textbf{.52} \\  
\bottomrule
\end{tabular}%
\caption{Applying LVLM-aware multimodal retrieval fine-tuning consistently boosts performance on 
inconsistent retrieval predictions across all datasets.
}
\label{tab:retrieval_alignment_training}
\end{table*}


\begin{table}
\centering
\scriptsize
\setlength{\tabcolsep}{1mm} 

\begin{tabular}{@{}lcccccccc@{}}
\toprule
Model & Breast & Derma & Retina & VinDr & BRSET & VQARAD & SLAKE   \\

\midrule
Qwen2-vl & 3\% & 51\% & 15\% & 50\% & 93\% & 12\% & 16\%  \\
\midrule
Pixtral  & 16\% & 13\% & 16\% & 66\% & 70\% & 7\% & 8\%   \\
\bottomrule
\end{tabular}%
\footnotesize
\caption{Proportion of inconsistent prediction cases.}
\label{tab:inconsistent_prediction_table}
\end{table}

{\textbf{Benchmarking versus medically pretrained LVLMs for classification.}
In Table~\ref{tab:binary_multi_class_table}, we include previously reported results for leading medically pretrained LVLM methods (detailed in Section~\ref{setup}). \methodname{} demonstrates results competitive with models that underwent extensive medical pretraining (pretraining sizes in Table~\ref{tab:medical_pretraining}). \methodname{} with the Pixtral backbone matches or exceeds all models except GSCo on the Breast dataset. To our knowledge, no medically pretrained LVLMs have been evaluated on the Retina benchmark, and we are not aware of any prior LVLM results on our other classification benchmarks. For binary and multi-class classification, we report only accuracy (ACC), as prior works did not report F1 scores. Conversely, for multi-label classification, we report only F1 scores. We also evaluated a medical baseline, Med-Flamingo, which generally underperformed and showed limited benefit from retrieval augmentation. We attribute this to its relatively older LLaMA backbone~\cite{touvron2023llama}. Additionally, we experimented with using BiomedCLIP~\cite{johnson2019billion} as the retriever for our general-purpose LVLMs, which exhibited a similar trend. Full results in Appendix~\ref{medflamingo}.}

\textbf{Benchmarking against pretrained LVLMs for visual question answering.} In Table~\ref{tab:compare_vqa}, we compare our approach with medically pretrained LVLM methods (detailed in Section~\ref{setup}) for VQA. \methodname{} achieves competitive performance relative to extensively pretrained models. For closed-question tasks, \methodname{} outperforms or matches existing models. For open-ended question answering, \methodname{} with the Qwen2-vl backbone matches LLaVA-Med\textsubscript{Vicuna} on the SLAKE dataset and LLaVA-Med\textsubscript{LLaVA} on the PathVQA dataset. Note that we do not report results on VQA-RAD, since our method was trained using an internal split of training and validation, whereas VQA-RAD provides only official training and test splits.

\section{Analysis}
\label{sec:analysis}


\textbf{\methodname{} boosts empirically challenging cases.} We observe that the performance gap is substantially larger for predictions classified as inconsistent retrieval predictions (Table \ref{tab:inconsistent_prediction_table} shows the proportion of such cases). We define an inconsistent retrieval prediction as one in which the model predicts different labels for retrieved candidates given the same query (see Figure \ref{fig:motivation}), i.e., if the model's predictions vary across retrieved candidates. Conversely, if the model predicts the same label for all retrieved candidates, the instance is consistent. {For testing the performance of the non-retrieval model, we feed the model only the target image, classified as consistent or inconsistent by the RAG model, without the retrieved candidates.}

Consistent and inconsistent prediction sets are constructed as follows: after the initial reader 
training stage, we run inference on the validation set and group instances by whether the reader 
produces consistent or inconsistent predictions across their retrieved candidates (as defined above). 
Performance on these two fixed subsets is then assessed after the LVLM-aware multimodal retrieval 
fine-tuning stage, allowing a direct comparison of how the fine-tuning affects each group. Detailed results are presented in Table \ref{tab:retrieval_alignment_training}. Overall, \methodname{} improves performance on inconsistent retrieval predictions by +0.12 in accuracy and +0.13 in F1 score when using the Qwen2‑vl backbone, and by +0.09 in both accuracy and F1 score with the Pixtral backbone, while also offering a slight improvement on the consistent. We observed that inconsistent retrieval predictions are empirically more challenging for the reader even without retrieval, FT RAG and \methodname{}, with a 10–20 point performance gap. 


{We hypothesize that our method improves inconsistent retrieval cases because conflicting predictions often signal misleading candidates. By aligning retriever scores with LVLM confidence via KL divergence, we upweight helpful candidates and suppress misleading ones. This is supported by GPT-5.2 relevance judgments, showing a 17\% improvement in retrieval quality (as described next). }

\begin{table*}
\centering
\footnotesize
\setlength{\tabcolsep}{1mm} 
\begin{tabular}{@{}lc|c|c|c|c|c|c|c|c|c@{}}
\toprule
Backbone   & Winner & Breast & Derma & Retina & VinDr & BRSET & VQA-RAD & SLAKE & PathVQA & \textbf{Mean}  \\
\midrule 
Qwen2-vl   & Tie              & .20 & .59 & .23 & .25 & .50 & .24 & .50 & .20 & .34 \\  
           & Base retriever Wins      & .30 & .09 & .32 & .36 & .24 & .25 & .16 & .35 & .26 \\ 
           & CLARE-retriever Wins & .50 & .30 & .44 & .46 & .25 & .50 & .34 & .45 & .41 \\\cmidrule(lr){2-11}
Pixtral    & Tie              & .24 & .51 & .51 & .47 & .35 & .12 & .15 & .29 & .33 \\  
           & Base retriever Wins     & .26 & .12 & .15 & .19 & .40 & .32 & .37 & .30 & .26 \\ 
           & CLARE-retriever Wins & .50 & .35 & .34 & .34 & .52 & .54 & .47 & .41 & .43 \\
\bottomrule
\end{tabular}%
\caption{Analysis of relevance of retrieval candidates before and after LVLM-aware multimodal retrieval fine-tuning. Training the multimodal retrieval significantly improves performance.}
\label{tab:relevant_retrieval_candidates}
\end{table*}

\begin{table}[t!]
\setlength{\tabcolsep}{0.8pt}
\scriptsize
\centering
\begin{tabular}{@{}lcccccc@{}}
\toprule
Backbone & Retrieval type & Breast & Derma & Retina & VQA-RAD & Mean \\
         &                & F1     & F1    & F1     & ACC     & F1   \\ \midrule
Flip a coin & & .30 & .10 & .16 & .50 & .31 \\ \midrule
& W/O Query img        & .42 & .18 & .33 & .67 & .49 \\
Pixtral & Reader only    & .66 & .53 & .44 & .72 & .65 \\
& W/O Retrieval    & .65 & .55 & .48 & .72 & .65 \\
& Random Retrieval & .69 & .56 & .45 & .70 & .65 \\ \midrule
& W/O Query img        & .52 & .20 & .20 & .70 & .48 \\
Qwen2-vl & Reader only   & .63 & .27 & .44 & .74 & .59 \\
& WO Retrieval    & .70 & .33 & .42 & .76 & .62 \\
& Random Retrieval & .67 & .39 & .42 & .77 & .62 \\
\bottomrule
\end{tabular}%
\footnotesize
\caption{Robustness evaluation. Despite an obvious drop in the extreme scenario of removing the query image, CLARE shows relative robustness.} 
\label{tab:model_robustness}
\end{table}

{\paragraph{Retrieval Relevance Analysis} We explore whether retrieval fine-tuning improves the quality of retrieval candidates compared to those produced before fine-tuning. Importantly, our setup does not include ground-truth retrieval labels; instead, the retriever learns to prioritize candidates that better steer the frozen LVLM toward the correct downstream prediction. To evaluate candidate improvement, we compare the retriever’s image and text heads before and after fine-tuning in a head-to-head ranking task. We then prompt GPT-5.2 to judge which candidate is more relevant to the ground-truth prediction, under the assumption that higher relevancy to the ground-truth prediction reflects stronger evidence for guiding the model. When both candidates are equally relevant, we record a tie. To mitigate position bias, we randomly shuffle candidate order before presenting them to the model. To reduce computational cost, we conduct the evaluation on a subset comprising 10\% of the samples. Table \ref{tab:relevant_retrieval_candidates} shows that retrieval fine-tuning improves candidate relevance, with a mean gain of 0.17 for both Qwen2-VL and Pixtral.}


{\paragraph{Retrieval Robustness Analyses.} We evaluate the robustness of \methodname{} under two retrieval-degraded conditions. First, we replace retrieved candidates with random ones. As shown in Table \ref{tab:model_robustness}, performance remains above the reader-only baseline across all datasets, indicating the model retains utility even under highly noisy retrieval. Second, we remove retrieved context entirely to assess whether the model preserves its intrinsic reader capabilities, a desirable property for deployment scenarios where no relevant candidates are found. In this setting, \methodname{} achieves performance comparable to, or slightly below, the reader-only model, confirming that retrieval fine-tuning does not degrade the model's standalone ability.}

{We further probe the informational value of retrieved candidates by \textbf{removing the input image entirely}, forcing the model to rely solely on retrieved (image, report/caption) pairs. Since the model no longer has direct access to the patient's condition, a substantial performance drop was expected. Using random guessing as a lower-bound baseline, \methodname{} exceeded it by a large margin (+0.32 in Accuracy, +0.28 in F1), demonstrating that the retrieved data carries meaningful predictive signal. Taken together, these analyses confirm that \methodname{} effectively leverages retrieved information without becoming overly dependent on it.}

\textbf{Performance with reranking.}
\label{rerank_analysis}
{Results for inconsistent predictions remain relatively low (Table \ref{tab:retrieval_alignment_training}). However, in the oracle analysis reported above, we found that \methodname{} can sometimes retrieve images that lead to correct predictions, together with other images that lead to wrong predictions. We thus explore whether a state-of-the-art model, when used as an optional reranker, could close the gap toward oracle performance. Specifically, we investigated whether the o3 model\cite{o32025openai} could select a retrieved image that leads to correct prediction. We provided o3 with an image to classify, along with four retrieved pairs, and tasked it with identifying the one containing the most predictive information. } Our results show that o3's performance generally surpasses simply taking the image with highest confidence, but o3 is generally inferior compared to fusing logits in \methodname{}. We conclude that reranking in this setting remains a significant challenge.  Results \Appendix \ref{rerank_analysis_a}.




\section{Conclusion}
\label{sec:conclusion}

We demonstrated that LVLM-aware multimodal retrieval achieves competitive medical diagnosis performance through lightweight, data-efficient fine-tuning, without medical pre-training. Future work may explore whether this approach can serve as a broadly applicable, cost-effective alternative to domain-specific pre-training when task-specific data and domain-specific knowledge bases are available but large-scale pre-training is prohibitive. 
In addition, we identified and substantially mitigated inconsistent retrieval predictions, where different retrieved candidates lead to conflicting diagnoses. Our oracle analysis revealed a considerable performance gap between actual results and what was theoretically achievable using the retrieved images, providing a foundation for future research aimed at narrowing it. Future work may also look into the prevalence and impact of inconsistent retrieval predictions more broadly.

\section*{Limitations}

Scope of evaluation. Our evaluation focuses on classification and visual question-answering tasks, including both open questions (a text-generation setting) and closed questions (closed-form answer selection). We did not evaluate our method on report generation, which requires different capabilities (longer-form generation, clinical writing conventions) and would benefit from dedicated investigation. Additionally, while our method shows consistent improvements across diverse imaging modalities (ultrasound, fundus photography, X-ray, histopathology, dermatoscopy, CT) and tasks (visual question answering, malignant lesion classification, diabetic retinopathy severity grading), evaluation on additional specialized modalities (e.g., PET, Microscopy, OCT) and tasks (e.g., organ classification, retinal OCT disease classification) would further validate generalizability.




\bibliography{custom}

@inproceedings{sun2024surf,
  title={SURf: Teaching large vision-language models to selectively utilize retrieved information},
  author={Sun, Jiashuo and Zhang, Jihai and Zhou, Yucheng and Su, Zhaochen and Qu, Xiaoye and Cheng, Yu},
  booktitle={Proceedings of the 2024 Conference on Empirical Methods in Natural Language Processing},
  pages={7611--7629},
  year={2024}
}

@article{he2024meddr,
  title={Meddr: Diagnosis-guided bootstrapping for large-scale medical vision-language learning},
  author={He, Sunan and Nie, Yuxiang and Chen, Zhixuan and Cai, Zhiyuan and Wang, Hongmei and Yang, Shu and Chen, Hao},
  journal={CoRR},
  year={2024}
}

@article{he2024gsco,
  title={Gsco: Towards generalizable ai in medicine via generalist-specialist collaboration},
  author={He, Sunan and Nie, Yuxiang and Wang, Hongmei and Yang, Shu and Wang, Yihui and Cai, Zhiyuan and Chen, Zhixuan and Xu, Yingxue and Luo, Luyang and Xiang, Huiling and others},
  journal={arXiv preprint arXiv:2404.15127},
  year={2024}
}

@article{izacard2023atlas,
  title={Atlas: Few-shot learning with retrieval augmented language models},
  author={Izacard, Gautier and Lewis, Patrick and Lomeli, Maria and Hosseini, Lucas and Petroni, Fabio and Schick, Timo and Dwivedi-Yu, Jane and Joulin, Armand and Riedel, Sebastian and Grave, Edouard},
  journal={Journal of Machine Learning Research},
  volume={24},
  number={251},
  pages={1--43},
  year={2023}
}

@inproceedings{hu2023reveal,
  title={Reveal: Retrieval-augmented visual-language pre-training with multi-source multimodal knowledge memory},
  author={Hu, Ziniu and Iscen, Ahmet and Sun, Chen and Wang, Zirui and Chang, Kai-Wei and Sun, Yizhou and Schmid, Cordelia and Ross, David A and Fathi, Alireza},
  booktitle={Proceedings of the IEEE/CVF conference on computer vision and pattern recognition},
  pages={23369--23379},
  year={2023}
}

@inproceedings{lin2023ra,
  title={Ra-dit: Retrieval-augmented dual instruction tuning},
  author={Lin, Xi Victoria and Chen, Xilun and Chen, Mingda and Shi, Weijia and Lomeli, Maria and James, Richard and Rodriguez, Pedro and Kahn, Jacob and Szilvasy, Gergely and Lewis, Mike and others},
  booktitle={The Twelfth International Conference on Learning Representations},
  year={2024}
}

@article{siriwardhana2023improving,
  title={Improving the domain adaptation of retrieval augmented generation (RAG) models for open domain question answering},
  author={Siriwardhana, Shamane and Weerasekera, Rivindu and Wen, Elliott and Kaluarachchi, Tharindu and Rana, Rajib and Nanayakkara, Suranga},
  journal={Transactions of the Association for Computational Linguistics},
  volume={11},
  pages={1--17},
  year={2023},
  publisher={MIT Press One Broadway, 12th Floor, Cambridge, Massachusetts 02142, USA~…}
}

@misc{epic2022,
    author ={Casey, Ross},
    url={https://www.statnews.com/2022/10/03/epic-sepsis-algorithm-revamp-training/},
    year = {2022}
}

@inproceedings{yuan2023ramm,
  title={Ramm: Retrieval-augmented biomedical visual question answering with multi-modal pre-training},
  author={Yuan, Zheng and Jin, Qiao and Tan, Chuanqi and Zhao, Zhengyun and Yuan, Hongyi and Huang, Fei and Huang, Songfang},
  booktitle={Proceedings of the 31st ACM International Conference on Multimedia},
  pages={547--556},
  year={2023}
}

@inproceedings{xia2024rule,
  title={Rule: Reliable multimodal rag for factuality in medical vision language models},
  author={Xia, Peng and Zhu, Kangyu and Li, Haoran and Zhu, Hongtu and Li, Yun and Li, Gang and Zhang, Linjun and Yao, Huaxiu},
  booktitle={Proceedings of the 2024 Conference on Empirical Methods in Natural Language Processing},
  pages={1081--1093},
  year={2024}
}

@article{xia2024mmed,
  title={Mmed-rag: Versatile multimodal rag system for medical vision language models},
  author={Xia, Peng and Zhu, Kangyu and Li, Haoran and Wang, Tianze and Shi, Weijia and Wang, Sheng and Zhang, Linjun and Zou, James and Yao, Huaxiu},
  journal={arXiv preprint arXiv:2410.13085},
  year={2024}
}

@article{wu2025mkgf,
  title={MKGF: A multi-modal knowledge graph based RAG framework to enhance LVLMs for Medical visual question answering},
  author={Wu, Yinan and Lu, Yuming and Zhou, Yan and Ding, Yifan and Liu, Jingping and Ruan, Tong},
  journal={Neurocomputing},
  pages={129999},
  year={2025},
  publisher={Elsevier}
}

@article{shi2023replug,
  title={Replug: Retrieval-augmented black-box language models},
  author={Shi, Weijia and Min, Sewon and Yasunaga, Michihiro and Seo, Minjoon and James, Rich and Lewis, Mike and Zettlemoyer, Luke and Yih, Wen-tau},
  journal={arXiv preprint arXiv:2301.12652},
  year={2023}
}

@inproceedings{thawakar2024xraygpt,
  title={XrayGPT: Chest radiographs summarization using large medical vision-language models},
  author={Thawakar, Omkar Chakradhar and Shaker, Abdelrahman M and Mullappilly, Sahal Shaji and Cholakkal, Hisham and Anwer, Rao Muhammad and Khan, Salman and Laaksonen, Jorma and Khan, Fahad},
  booktitle={Proceedings of the 23rd workshop on biomedical natural language processing},
  pages={440--448},
  year={2024}
}

@article{jeong2024limited,
  title={The Limited Impact of Medical Adaptation of Large Language and Vision-Language Models},
  author={Jeong, Daniel P and Mani, Pranav and Garg, Saurabh and Lipton, Zachary C and Oberst, Michael},
  journal={arXiv preprint arXiv:2411.08870},
  year={2024}
}

@article{tschandl2018ham10000,
  title={The HAM10000 dataset, a large collection of multi-source dermatoscopic images of common pigmented skin lesions},
  author={Tschandl, Philipp and Rosendahl, Cliff and Kittler, Harald},
  journal={Scientific data},
  volume={5},
  number={1},
  pages={1--9},
  year={2018},
  publisher={Nature Publishing Group}
}

@article{liu2022deepdrid,
  title={Deepdrid: Diabetic retinopathy—grading and image quality estimation challenge},
  author={Liu, Ruhan and Wang, Xiangning and Wu, Qiang and Dai, Ling and Fang, Xi and Yan, Tao and Son, Jaemin and Tang, Shiqi and Li, Jiang and Gao, Zijian and others},
  journal={Patterns},
  volume={3},
  number={6},
  year={2022},
  publisher={Elsevier}
}

@inproceedings{chen2024internvl,
  title={Internvl: Scaling up vision foundation models and aligning for generic visual-linguistic tasks},
  author={Chen, Zhe and Wu, Jiannan and Wang, Wenhai and Su, Weijie and Chen, Guo and Xing, Sen and Zhong, Muyan and Zhang, Qinglong and Zhu, Xizhou and Lu, Lewei and others},
  booktitle={Proceedings of the IEEE/CVF conference on computer vision and pattern recognition},
  pages={24185--24198},
  year={2024}
}

@article{wu2023towards,
  title={Towards generalist foundation model for radiology by leveraging web-scale 2D\&3D medical data},
  author={Wu, Chaoyi and Zhang, Xiaoman and Zhang, Ya and Wang, Yanfeng and Xie, Weidi},
  journal={arXiv preprint arXiv:2308.02463},
  year={2023}
}

@article{zhang2023pmc,
  title={Pmc-vqa: Visual instruction tuning for medical visual question answering},
  author={Zhang, Xiaoman and Wu, Chaoyi and Zhao, Ziheng and Lin, Weixiong and Zhang, Ya and Wang, Yanfeng and Xie, Weidi},
  journal={arXiv preprint arXiv:2305.10415},
  year={2023}
}

@inproceedings{moor2023med,
  title={Med-flamingo: a multimodal medical few-shot learner},
  author={Moor, Michael and Huang, Qian and Wu, Shirley and Yasunaga, Michihiro and Dalmia, Yash and Leskovec, Jure and Zakka, Cyril and Reis, Eduardo Pontes and Rajpurkar, Pranav},
  booktitle={Machine Learning for Health (ML4H)},
  pages={353--367},
  year={2023},
  organization={PMLR}
}

@article{li2023llava,
  title={Llava-med: Training a large language-and-vision assistant for biomedicine in one day},
  author={Li, Chunyuan and Wong, Cliff and Zhang, Sheng and Usuyama, Naoto and Liu, Haotian and Yang, Jianwei and Naumann, Tristan and Poon, Hoifung and Gao, Jianfeng},
  journal={Advances in Neural Information Processing Systems},
  volume={36},
  pages={28541--28564},
  year={2023}
}

@article{wang2024qwen2,
  title={Qwen2-vl: Enhancing vision-language model's perception of the world at any resolution},
  author={Wang, Peng and Bai, Shuai and Tan, Sinan and Wang, Shijie and Fan, Zhihao and Bai, Jinze and Chen, Keqin and Liu, Xuejing and Wang, Jialin and Ge, Wenbin and others},
  journal={arXiv preprint arXiv:2409.12191},
  year={2024}
}

@article{agrawal2024pixtral,
  title={Pixtral 12B},
  author={Agrawal, Pravesh and Antoniak, Szymon and Hanna, Emma Bou and Bout, Baptiste and Chaplot, Devendra and Chudnovsky, Jessica and Costa, Diogo and De Monicault, Baudouin and Garg, Saurabh and Gervet, Theophile and others},
  journal={arXiv preprint arXiv:2410.07073},
  year={2024}
}

@article{yoran2023making,
  title={Making retrieval-augmented language models robust to irrelevant context},
  author={Yoran, Ori and Wolfson, Tomer and Ram, Ori and Berant, Jonathan},
  journal={arXiv preprint arXiv:2310.01558},
  year={2023}
}

@article{al2020dataset,
  title={Dataset of breast ultrasound images},
  author={Al-Dhabyani, Walid and Gomaa, Mohammed and Khaled, Hussien and Fahmy, Aly},
  journal={Data in brief},
  volume={28},
  pages={104863},
  year={2020},
  publisher={Elsevier}
}

@article{lau2018dataset,
    title={A dataset of clinically generated visual questions and answers about radiology images},
    author={Lau, Jason J and Gayen, Soumya and Ben Abacha, Asma and Demner-Fushman, Dina},
    journal={Scientific data},
    volume={5},
    number={1},
    pages={1--10},
    year={2018},
    publisher={Nature Publishing Group}
}

@inproceedings{lin2023pmc,
  title={Pmc-clip: Contrastive language-image pre-training using biomedical documents},
  author={Lin, Weixiong and Zhao, Ziheng and Zhang, Xiaoman and Wu, Chaoyi and Zhang, Ya and Wang, Yanfeng and Xie, Weidi},
  booktitle={International Conference on Medical Image Computing and Computer-Assisted Intervention},
  pages={525--536},
  year={2023},
  organization={Springer}
}

@article{johnson2019mimic,
  title={MIMIC-CXR, a de-identified publicly available database of chest radiographs with free-text reports},
  author={Johnson, Alistair EW and Pollard, Tom J and Berkowitz, Seth J and Greenbaum, Nathaniel R and Lungren, Matthew P and Deng, Chih-ying and Mark, Roger G and Horng, Steven},
  journal={Scientific data},
  volume={6},
  number={1},
  pages={317},
  year={2019},
  publisher={Nature Publishing Group UK London}
}

@article{ruckert2024rocov2,
  title={Rocov2: Radiology objects in context version 2, an updated multimodal image dataset},
  author={R{\"u}ckert, Johannes and Bloch, Louise and Br{\"u}ngel, Raphael and Idrissi-Yaghir, Ahmad and Sch{\"a}fer, Henning and Schmidt, Cynthia S and Koitka, Sven and Pelka, Obioma and Abacha, Asma Ben and G. Seco de Herrera, Alba and others},
  journal={Scientific Data},
  volume={11},
  number={1},
  pages={688},
  year={2024},
  publisher={Nature Publishing Group UK London}
}

@article{medmnistv2,
    title={MedMNIST v2-A large-scale lightweight benchmark for 2D and 3D biomedical image classification},
    author={Yang, Jiancheng and Shi, Rui and Wei, Donglai and Liu, Zequan and Zhao, Lin and Ke, Bilian and Pfister, Hanspeter and Ni, Bingbing},
    journal={Scientific Data},
    volume={10},
    number={1},
    pages={41},
    year={2023},
    publisher={Nature Publishing Group UK London}
}

@article{zhang2023biomedclip,
  title={Biomedclip: a multimodal biomedical foundation model pretrained from fifteen million scientific image-text pairs},
  author={Zhang, Sheng and Xu, Yanbo and Usuyama, Naoto and Xu, Hanwen and Bagga, Jaspreet and Tinn, Robert and Preston, Sam and Rao, Rajesh and Wei, Mu and Valluri, Naveen and others},
  journal={arXiv preprint arXiv:2303.00915},
  year={2023}
}

@article{johnson2019billion,
  title={Billion-scale similarity search with {GPUs}},
  author={Johnson, Jeff and Douze, Matthijs and J{\'e}gou, Herv{\'e}},
  journal={IEEE Transactions on Big Data},
  volume={7},
  number={3},
  pages={535--547},
  year={2019},
  publisher={IEEE}
}

@article{lambert2022trustworthy,
  title={Trustworthy clinical AI solutions: a unified review of uncertainty quantification in deep learning models for medical image analysis},
  author={Lambert, Benjamin and Forbes, Florence and Tucholka, Alan and Doyle, Senan and Dehaene, Harmonie and Dojat, Michel},
  journal={arXiv preprint arXiv:2210.03736},
  year={2022}
}

@inproceedings{zheng2024llamafactory,
  title={LlamaFactory: Unified Efficient Fine-Tuning of 100+ Language Models},
  author={Yaowei Zheng and Richong Zhang and Junhao Zhang and Yanhan Ye and Zheyan Luo and Zhangchi Feng and Yongqiang Ma},
  booktitle={Proceedings of the 62nd Annual Meeting of the Association for Computational Linguistics (Volume 3: System Demonstrations)},
  address={Bangkok, Thailand},
  publisher={Association for Computational Linguistics},
  year={2024},
  url={http://arxiv.org/abs/2403.13372}
}

@misc{hu2021loralowrankadaptationlarge,
      title={LoRA: Low-Rank Adaptation of Large Language Models}, 
      author={Edward J. Hu and Yelong Shen and Phillip Wallis and Zeyuan Allen-Zhu and Yuanzhi Li and Shean Wang and Lu Wang and Weizhu Chen},
      year={2021},
      eprint={2106.09685},
      archivePrefix={arXiv},
      primaryClass={cs.CL},
      url={https://arxiv.org/abs/2106.09685}, 
}

@article{pham2022vindr,
  title={VinDr-PCXR: An open, large-scale pediatric chest X-ray dataset for interpretation of common thoracic diseases},
  author={Pham, H Hieu and Tran, T Thanh and Nguyen, Ha Quy},
  journal={PhysioNet (version 1.0. 0)},
  volume={10},
  number={2},
  year={2022}
}

@article{o32025openai,
  title={OpenAI o3 and o4-mini System Card},
  author={OpenAI},
  url={https://cdn.openai.com/pdf/2221c875-02dc-4789-800b-e7758f3722c1/o3-and-o4-mini-system-card.pdf},
  year={2025}
}

@article{he2020pathvqa,
  title={Pathvqa: 30000+ questions for medical visual question answering},
  author={He, Xuehai and Zhang, Yichen and Mou, Luntian and Xing, Eric and Xie, Pengtao},
  journal={arXiv preprint arXiv:2003.10286},
  year={2020}
}

@article{luo2023biomedgpt,
  title={Biomedgpt: Open multimodal generative pre-trained transformer for biomedicine},
  author={Luo, Yizhen and Zhang, Jiahuan and Fan, Siqi and Yang, Kai and Wu, Yushuai and Qiao, Mu and Nie, Zaiqing},
  journal={arXiv preprint arXiv:2308.09442},
  year={2023}
}

@article{izacard2020leveraging,
  title={Leveraging passage retrieval with generative models for open domain question answering},
  author={Izacard, Gautier and Grave, Edouard},
  journal={arXiv preprint arXiv:2007.01282},
  year={2020}
}

@inproceedings{xiao2024jina,
  title={Jina CLIP: Your CLIP model is also your text retriever},
  author={Xiao, Han and Mastrapas, Georgios and Wang, Bo},
  booktitle={Multi-modal Foundation Model meets Embodied AI Workshop@ ICML2024},
  year={2024}
}

@inproceedings{liu2021slake,
  title={Slake: A semantically-labeled knowledge-enhanced dataset for medical visual question answering},
  author={Liu, Bo and Zhan, Li-Ming and Xu, Li and Ma, Lin and Yang, Yan and Wu, Xiao-Ming},
  booktitle={2021 IEEE 18th international symposium on biomedical imaging (ISBI)},
  pages={1650--1654},
  year={2021},
  organization={IEEE}
}

@article{touvron2023llama,
  title={Llama: Open and efficient foundation language models},
  author={Touvron, Hugo and Lavril, Thibaut and Izacard, Gautier and Martinet, Xavier and Lachaux, Marie-Anne and Lacroix, Timoth{\'e}e and Rozi{\`e}re, Baptiste and Goyal, Naman and Hambro, Eric and Azhar, Faisal and others},
  journal={arXiv preprint arXiv:2302.13971},
  year={2023}
}

@article{nakayama2024brazilian,
  title={A brazilian multilabel ophthalmological dataset (BRSET). 2023},
  author={Nakayama, LF and Goncalves, M and Zago Ribeiro, L and Santos, H and Ferraz, D and Malerbi, F and others},
  journal={URL: https://physionet. org/content/brazilian-ophthalmological/1.0. 0/[accessed 2024-08-14]},
  year={2024}
}

\appendix

\begin{table*}[t!]
\footnotesize
\begin{subtable}{\textwidth}
\centering
\caption{}
\label{tab:full_results_classification_analysis}
\begin{tabular}{@{}lc|cc|cc|cc|cc|cc|cc@{}}
\toprule
Backbone (Size) &  Model & \multicolumn{2}{c}{Breast} & \multicolumn{2}{c}{Derma} & \multicolumn{2}{c}{Retina} & \multicolumn{2}{c}{VinDr-PCXR} & \multicolumn{2}{c}{BRSET} & \multicolumn{2}{c}{Mean} \\
\cmidrule(lr){3-4} \cmidrule(lr){5-6} \cmidrule(lr){7-8} \cmidrule(lr){9-10} \cmidrule(lr){11-12} \cmidrule(lr){13-14}
& & ACC & F1 & ACC & F1 & ACC & F1  & ACC & F1 & ACC & F1 & ACC & F1 \\ 
\midrule 
Qwen-vl (7B)       &   FT RAG & .85 & .82 & .71 & .42 & .62 & .48 & .55 & .09 & .48 & .27 & .64 & .42 \\ \cmidrule(lr){2-14}
&   Text Retriever Head Only           & .86 & .83 & .74 & .46 & .62 & .48 & .56 & .13 & .48 & .32 & .65 & .44   \\ \cmidrule(lr){2-14}
&   Image Retriever Head Only           & .87 & .83 & .75 & .48 & .64 & .49 & .56 & .13 & .48 & .30 & .66 & .45   \\ \cmidrule(lr){2-14}
          &    \methodname{}  & \textbf{.87} & \textbf{.84} & \textbf{.76} & \textbf{.50} & \textbf{.65} & \textbf{.50} & \textbf{.57} & \textbf{.14} & \textbf{.49} & \textbf{.37} & \textbf{.67} & \textbf{.47} \\ \midrule     
      
Pixtral (12B) &    FT RAG & .88 & .85 & .79 & .60 & .57 & .47 & .49 & .09 & .47 & .33 & .64 & .47 \\ \cmidrule(lr){2-14}
&          Text Retriever Head Only & .89 & .86 & .79 & .61 & .58 & .48 & .55 & .13 & .50 & .35 & .66 & .49 \\ \cmidrule(lr){2-14}
&          Text Retriever Head Only & .90 & .86 & .79 & .60 & .59 & .48 & .54 & .13 & .49 & .35 & .66 & .48 \\ \cmidrule(lr){2-14}
      &    \methodname{}  & \textbf{.90} & \textbf{.87} & \textbf{.80} & \textbf{.62} & \textbf{.60} & \textbf{.51} & \textbf{.56} & \textbf{.14} & \textbf{.51} & \textbf{.37} & \textbf{.67} & \textbf{.50} \\ 
         
\bottomrule
\end{tabular}
\end{subtable}
\begin{subtable}{\textwidth}
\centering
\caption{}
\label{tab:full_results_vqa_analysis}
\begin{tabular}{@{}lc|cc|cc|cc|cc@{}}
\toprule
Backbone (Size) & Model & \multicolumn{2}{c}{VQA-RAD} & \multicolumn{2}{c}{SLAKE} & \multicolumn{2}{c}{PathVQA} & \multicolumn{2}{c}{Mean} \\
\cmidrule(lr){3-4} \cmidrule(lr){5-6} \cmidrule(lr){7-8} \cmidrule(lr){9-10}
& & Closed & Open & Closed & Open & Closed & Open & Closed & Open   \\ \midrule
Qwen2-vl (7B) &     FT RAG     & .76 & .45 & .88 & .81 & .91 & .33 & .85 & .53 \\
        &     Text Retriever Head Only & .77 & .46 & .89 & .82 & \textbf{.93} & .35 & .86 & .54 \\
        &     Image Retriever Head Only & .78 & .47 & .89 & .82 & .92 & .34 & .86 & .54 \\
  &     Closed Question Only   & .78 & .47 & \textbf{.90} & .83 & \textbf{.93} & .37 & \textbf{.87} & .56 \\ \cmidrule(lr){2-10}
         &    \methodname{}    & \textbf{.79} & \textbf{.48} & \textbf{.90} & \textbf{.84} & \textbf{.93} & \textbf{.38} & \textbf{.87} & \textbf{.57}    \\ \midrule
         
Pixtral (12B)  &    FT RAG  & .74 & .41 & .88 & .81 & .88 & .31 & .83 & .51  \\
     &     Text Retriever Head Only & .76 & .44 & \textbf{.90} & \textbf{.83} & .89 & .33 & .85 & .53 \\
     &     Image Retriever Head Only& .76 & .44 & .89 & .82 & .90 & .32 & .85 & .53 \\
&     Closed Question Only  & .76 & .45 & \textbf{.90} & \textbf{.84} & .90 & .36 & .85 & .55  \\ \cmidrule(lr){2-10}
&     \methodname{}         & \textbf{.78} & \textbf{.47} & \textbf{.90} & \textbf{.83} & \textbf{.93} & \textbf{.37} & \textbf{.87} & \textbf{.56} \\ 
         
\bottomrule
\end{tabular}
\end{subtable}
\caption{Ablation results for classification (a) and visual question answering (b). We demonstrate the necessity of each stage in our training paradigm. The first stage, FT RAG, trains only the reader. Subsequently, training the text retriever improves performance, and finally, completing the process with the image retriever (\methodname{}) yields the best results. We also evaluate training only on closed questions—without applying the \texttt{o3} model to convert open questions into closed ones—and observe a gain in open-question performance in this setting 
}
\end{table*}

\section{Dataset}
\label{dataset_all}

\subsection{Dataset for Evaluation}
\label{eval_datasets}

Our datasets include real-world hospital datasets (BRSET \cite{nakayama2024brazilian} and VinDr‑PCXR \cite{pham2022vindr}) alongside a variety of classification and visual question answering datasets. We focus on a low-resource data-efficient setting (training sets ranging from 546 to 7007 samples in classification, 1790-19,700 in VQA). Medical image annotation is a resource-intensive task, demanding expert annotators whose availability is limited and expensive. Clinical AI also often has poor generalization across heterogeneous medical centers and patient populations, often requiring that models be trained for the specific context in which they are deployed \cite{epic2022}.
Such constraints often restrict researchers to datasets comprising only a few thousand samples for a given study. Retrieval augmentation is well-suited for this setting, as it is known to benefit low‑data regimes the most by leveraging external knowledge to compensate for sparse training samples. For the classification tasks, we used the following datasets:

BreastMNIST (nickname: Breast) \cite{al2020dataset} is a binary classification dataset of breast ultrasound. Following the official split, we use 546 for training, 78 for validation and 156 for testing.

RetinaMNIST (nickname: Retina) \cite{liu2022deepdrid} is a multi-label classification dataset of retina Fundus Camera. Following the official split, we use 1,080 for training, 120 for validation and 400 for testing.

DermaMNIST (nickname: Derma) \cite{tschandl2018ham10000} is a multi-class classification dataset of common pigmented skin lesions. Following the official split, we use 7,007 for training, 1,003 for validation and 2,005 for testing. The dataset contains clinical images from 7 different diagnostic categories: actinic keratoses, basal cell carcinoma, benign keratosis, dermatofibroma, melanoma, melanocytic nevi, and vascular lesions.

VinDr-PCXR \cite{pham2022vindr} is a large, open pediatric chest X-ray dataset collected in Vietnam (2020–2021) containing 9,125 posteroanterior scans from patients under 10 years old. It provides both lesion-level bounding-box annotations for 36 findings and image-level labels for multi-label of 15 diagnoses, curated by experienced radiologists. We follow the official split, which includes 7,728 training and 1,397 test images, with an additional internal split of the training data into 85\% for training and 15\% for validation.

BRSET \cite{nakayama2024brazilian} is a Brazilian multilabel ophthalmological dataset comprising retinal fundus images annotated with 14 distinct pathological findings. The dataset, composed of 16,266 images, presents a challenging real-world scenario for multilabel classification. There is no publicly available split; therefore, we apply an internal split for train, validation, and test sets (70\% train, 10\% validation, and 20\% test).

Breast, Derma, and Retina taken from the large-scale MNIST-like collection of standardized biomedical images, including 12 datasets for 2D and 6 datasets for 3D. We especially used MedMNIST+ \cite{medmnistv2}, which is a higher resolution extension of the original MedMNIST. We used the highest resolution of 224 × 224.

For medical question answering, we have three well-known datasets: 

VQA-RAD \cite{lau2018dataset} is a clinician-annotated visual question answering dataset focused on radiology images (e.g., X-ray, CT, MRI). It pairs each image with both open-ended and yes/no questions. Following the official split, we use 1,753 questions for training and 453 for testing. In addition, we further partition the training set into 85\% for training and 15\% for validation.

PathVQA \cite{he2020pathvqa} is a pathology-focused medical VQA dataset built from textbooks and the PEIR digital library, comprising 4,289 images and 32,632 question–answer pairs spanning both open-ended and yes/no questions. We follow the official split provided by the authors: 19,700 for training, 6,260 for validation, and 6,720 for testing.

SLAKE \cite{liu2021slake} is a bilingual medical VQA dataset. We use the English subset, \textit{SLAKE-English}, which comprises 642 radiology images and 7.03k English QA pairs. We follow the official train/validation/test split of 4.92k/1.05k/1.06k QA pairs, and include all question types, covering both open-ended and closed-ended formats.

\subsection{Dataset for Index}
\label{Index_datasets}
For construction of the Index we used three large datasets of (image, text) pairs: PMC-OA \cite{lin2023pmc}, ROCO \cite{ruckert2024rocov2} and MIMIC-CXR \cite{johnson2019mimic}:

PMC-OA: is a large-scale dataset that contains 1.65M image-text pairs. The figures and captions from PubMed Central, 2,478,267 available papers are covered.

ROCO: is an image-caption dataset collected from PubMed. It filters out all the compound or non-radiological images, and consists of 81K samples.

MIMIC-CXR: is the largest chest X-ray dataset, containing 377,110 samples (image-report pairs). Each image is paired with a clinical report describing findings from doctors.

\begin{table}[t!]
\centering
\footnotesize
\setlength{\tabcolsep}{1mm} 
\begin{tabular}{@{}lc|cc|cc|cc@{}}
\toprule
Backbone & Model & \multicolumn{2}{c}{Breast} & \multicolumn{2}{c}{Derma} & \multicolumn{2}{c}{Retina} \\
\cmidrule(lr){3-4} \cmidrule(lr){5-6} \cmidrule(lr){7-8}  
& &  ACC & F1 & ACC & F1 & ACC & F1   \\
\midrule
Qwen2-vl &   Top-1   &  .50  & .60   & .32   & .25   & .38   & .32 \\
        &  Top-1 logits         & .60  & .58   & .43 & .35   & .37   & .38  \\
         &  o3              &  .77  & .80   & .42   & .32   & .35   & .33 \\
        &  o3 multi-image &  .77  & .80   & .43   & .33   & .43   & .35 \\ 
         &  \methodname{}   &  .77  & .80   & .57   & .43   & .51   & .47 \\ \midrule
         
Pixtral  & Top-1 &   .72  & .75   & .46 & .48   & .29   & .33 \\
 &  Top-1 logits         &   .64  & .63   & .40 & .40  & .29 & .32 \\
         &   o3         &   .73  & .73   & .42 & .37   & .25   & .30  \\
&   o3  multi-image     &   .73  & .73       & .43 & .42   & .34   & .37  \\
& \methodname{}         &   .76  & .77   & .45 & .47   & .34   & .37 \\

\bottomrule
\end{tabular}
\caption{Study of o3 as a reranker. We evaluate o3's ability to rerank the inconsistent predictions, selecting which of the retrieved (image, caption) pairs has the most predictive information for the query image. We compared the performance of choosing the highest similarity retrieved (image, caption) pair, choosing the highest confidence (top logits), using only the caption for reranking, and using our current strategy of fusing logits from all predictions. The metrics used are accuracy and F1 score. }
\label{tab:rerankers}
\end{table}

\section{Implementation Details}
\label{implementation_details}
\subsection{Retrieval Augmentation}

As described earlier, for each image–question pair, we retrieve \(r\) image–text pairs, with \(r = 4\). We use the input image as the query and Jina-CLIP \cite{xiao2024jina} as the retriever head in our multimodal retriever. The index is constructed with FAISS \cite{johnson2019billion} over MIMIC-CXR, PMC-OA, and ROCO, which are fully described in Section~\ref{Index_datasets} \Appendix A. To embed images, we use the Jina-CLIP visual head (the same model used for retrieval); to embed captions/reports, we use the Jina-CLIP text head. The index is stored in \texttt{float16} due to storage constraints. We also experimented with BiomedCLIP \cite{zhang2023biomedclip}.

\subsection{Reader Fine-Tuning}
We use Pixtral (12B) \cite{agrawal2024pixtral} and Qwen2-vl (7B) \cite{wang2024qwen2} as base models. We train the models with and without retrieval augmentation to assess their effect. The retrieval-augmented prompt is described \Appendix E. Note, we also tried Med-Flamingo (9B) \cite{moor2023med} as a base model.

We fine-tune Pixtral and Qwen2-vl using LLaMA-Factory \cite{zheng2024llamafactory} on a single NVIDIA L40 (48\,GB) GPU for 10 epochs. We use a learning rate of \(2 \times 10^{-5}\) and apply Low-Rank Adaptation (LoRA) \cite{hu2021loralowrankadaptationlarge} for parameter-efficient fine-tuning. To all models (including baseline models),  we conduct a grid search over batch size (2, 4, 6) and whether to freeze the vision head, selecting the best configuration by validation performance. 
For Med-Flamingo, we fine-tune using our codebase on a single NVIDIA L40 (48\,GB) GPU for 10 epochs. Following the Med-Flamingo paper, the language model and image encoder are frozen, and only the Gated Cross-Attention layers and the Perceiver Resampler are optimized for stable and efficient learning. We use a learning rate of \(2 \times 10^{-5}\).

\subsection{LVLM-Aware Multimodal Retrieval Fine-Tuning}
We train the retrieval model to fetch relevant context while keeping the reader frozen. We use Jina-CLIP’s visual head as the base model for the multimodal retriever. The model is trained on a single NVIDIA L40 (48\,GB) GPU for 100 epochs with a learning rate of \(2 \times 10^{-5}\), freezing all layers except the last ten. We set the number of retrieved candidates per query to 50. In our experiments, retrieving more candidates further improved retriever performance.

\subsection{Evaluation Metrics}

\paragraph{Classification.} We report Accuracy (ACC) and Macro F1:
\[
\mathrm{ACC} = \frac{1}{N}\sum_{n=1}^{N}\mathbf{1}[\hat{y}_n = y_n], \qquad
\]

\[
\mathrm{MacroF1} = \frac{1}{C}\sum_{c=1}^{C}\frac{2\,\mathrm{Prec}_c\,\mathrm{Rec}_c}{\mathrm{Prec}_c+\mathrm{Rec}_c},
\]

where $\mathrm{Prec}_c$ and $\mathrm{Rec}_c$ are precision and recall computed per class $c$, $C$ is the number of classes, and $N$ is the number of examples.

\paragraph{Visual Question Answering (VQA).}
For open-ended questions, we compute token-level Recall and F1 between the predicted token set $\hat{A}$ and the reference token set $A$:
\[
\mathrm{Prec} = \frac{|\hat{A}\cap A|}{|\hat{A}|}, \quad
\mathrm{Rec} = \frac{|\hat{A}\cap A|}{|A|}, \quad
\]

\[
\mathrm{F1} = \frac{2\,\mathrm{Prec}\cdot \mathrm{Rec}}{\mathrm{Prec}+\mathrm{Rec}}.
\]
For closed-ended questions, we use Exact Match.

\begin{table*}[t!]
\centering
\setlength{\tabcolsep}{1mm} 
\begin{tabular}{@{}l|cc|cc|cc|cc|cc|c|c|c|cc@{}}
\toprule
\makecell{Model} & \multicolumn{2}{c}{Breast} & \multicolumn{2}{c}{Derma} & \multicolumn{2}{c}{Retina} & \multicolumn{2}{c}{Vindr} & \multicolumn{2}{c}{BREST} & \multicolumn{1}{c}{VQARAD} & \multicolumn{1}{c}{SLAKE} & \multicolumn{1}{c}{PathVQA} & \multicolumn{2}{c}{Mean} \\
\cmidrule(lr){2-3} \cmidrule(lr){4-5} \cmidrule(lr){6-7} \cmidrule(lr){8-9} \cmidrule(lr){10-11} \cmidrule(lr){12-12} \cmidrule(lr){13-13} \cmidrule(lr){14-14} \cmidrule(lr){15-16}
             & ACC & F1 & ACC & F1 & ACC & F1 & ACC & F1 & ACC & F1 & ACC & ACC & ACC & ACC & F1  \\ \midrule
 FT RAG\textsubscript{Qwen2-vl} 
            &.84 & .81 & .73 & .38 & .63 & .45  & .54 & .09 & .49 & .27 & .77 & .89 & \textbf{.92} & .72 & .40 \\
            \methodname{}\textsubscript{Qwen2-vl} &
            \textbf{.87} & \textbf{.82} & \textbf{.75} & \textbf{.40} & \textbf{.65} & \textbf{.50}  & \textbf{.58} & \textbf{.15} & \textbf{.50} & \textbf{.39} & \textbf{.79} & \textbf{.91} & \textbf{.92} & \textbf{.74} & \textbf{.45} \\ \midrule

            FT RAG\textsubscript{Pixtral} &
            .88 & .84 & .79 & .57 & .57 & .46  & .55 & .09 & .44 & .31 & .74 & .87 & .88 & .72 & .45\\
            \methodname{}\textsubscript{Pixtral} & 
            \textbf{.91} & \textbf{.87} & \textbf{.80} & \textbf{.59} & \textbf{.59} & \textbf{.49}  & \textbf{.56} & \textbf{.15} & \textbf{.50} & \textbf{.39} & \textbf{.77} & \textbf{.89} & \textbf{.89} & \textbf{.74} & \textbf{.50}\\ 
           
\bottomrule
\end{tabular}%
\footnotesize
\caption{Performance comparison of \methodname{} against FT RAG baseline across medical imaging classification and VQA tasks. Both Qwen2-VL and Pixtral backbones show consistent improvements with \methodname{}, particularly in F1 scores. Results demonstrate that BiomedCLIP-based retrieval enhances performance similarly to the general-purpose retriever. VQA results are reported for the closed-question subset using exact match accuracy.}
\label{tab:appendix_full_results_bio}
\end{table*}


\begin{table}[t!]
\centering
\setlength{\tabcolsep}{1mm} 
\begin{tabular}{@{}l|cc|cc|cc|c@{}}
\toprule
\makecell{Model} & \multicolumn{2}{c}{Breast} & \multicolumn{2}{c}{Derma} & \multicolumn{2}{c}{Retina} & VQARAD  \\
\cmidrule(lr){2-3} \cmidrule(lr){4-5} \cmidrule(lr){6-7} \cmidrule(lr){8-8} 
& ACC & F1 & ACC & F1 & ACC & F1 & ACC   \\ 
\midrule
\makecell{Reader\\only} & \textbf{.82} & \textbf{.77} & .67 & \textbf{.54} & .57 & .40 & .65   \\
\methodname{} & .81 & \textbf{.77} & \textbf{.68} & \textbf{.54} & \textbf{.60} & \textbf{.51} & \textbf{.69}   \\
\bottomrule
\end{tabular}%
\\[0.5em]
\footnotesize
\caption{Med-Flamingo comparison with and without \methodname{} retrieval augmentation across medical imaging tasks. \methodname{} achieves best performance on 4 out of 7 metrics, with notable improvements over Reader Only in Retina classification (ACC: 0.57→0.60, F1: 0.40→0.51) and VQA-RAD (0.65→0.69). However, the overall performance gains are more modest than those observed with other backbone architectures, indicating that Med-Flamingo may have limited capacity to leverage additional retrieval context.}
\label{tab:appendix_full_results}
\end{table}

\section{Baseline Methods}
\label{sec:baselines}

\subsection{RAG baseline}

We evaluate our method against several representative RAG-based and multimodal retrieval approaches to ensure a fair and comprehensive comparison.

\textbf{RAD} \cite{he2024meddr} is a retrieval-based method designed for classification tasks. During inference, the most similar image from the training set is retrieved, and its supervised label is directly used as the prediction. We integrate RAD with both of our backbone models (Qwen2-VL and Pixtral) under the same data splits and retrieval settings.

\textbf{MMed-RAG}\cite{xia2024mmed} represents a recent state-of-the-art multimodal RAG approach in which the retriever and LVLM are trained independently. We follow the official GitHub implementation, first training the CLIP module and then performing DPO training using two backbone models: LLaVA-Med and LLaVA.

\textbf{Fusion-in-Decoder (FiD) Pipeline} \cite{izacard2020leveraging}  serves as a standard RAG-style baseline. In this setting, the retriever is frozen, and only the reader is fine-tuned while processing retrieved image–caption pairs as contextual input.

\textbf{Perplexity Distillation Loss (PDist)} \cite{izacard2023atlas}. For each query and candidate document, we compute the reduction in perplexity (or equivalently, the increase in likelihood) of the correct output when the language model is conditioned on that document. From these likelihoods, we derive a “posterior” over documents (proportional to their contributions), and train the retriever to mimic that posterior via a KL divergence objective. In doing so, the retriever is guided to prioritize documents that most effectively support the language model’s generation, thereby aligning retrieval with generation quality.

\subsection{Medical Pre - trained baseline models}
We benchmark against state-of-the-art medical large vision--language models (LVLMs) that underwent large-scale medical pre-training:

\paragraph{BiomedGPT.}An open multimodal generative pre-trained transformer for biomedicine that aligns diverse biological/biomedical modalities with natural language; we include it as a strong domain baseline~\cite{luo2023biomedgpt}.

\paragraph{LLaVA-Med.}A biomedical adaptation of LLaVA trained via curriculum (biomedical figure--caption alignment followed by instruction tuning); we evaluate three commonly used releases as separate baselines~\cite{li2023llava}.

\paragraph{MedVInT-TE and MedVInT-TD.} Two implementations of Medical Visual Instruction Tuning from PMC-VQA~\cite{zhang2023pmc}. Both adopt a PMC-CLIP vision encoder~\cite{lin2023pmc}. TE uses an encoder-style text pathway feeding a multimodal decoder, while TD concatenates text tokens with visual features to a decoder-only pathway; we report both as distinct baselines.

\paragraph{InternVL-based baselines: MedDr, MedDr+RAD, and GSCo.} \textbf{MedDr} is a generalist medical VLLM trained on large-scale instruction-style data curated from diagnosis-guided bootstrapping and medical image descriptions, covering multiple modalities and tasks~\cite{he2024meddr}. 
\textbf{MedDr+RAD} augments MedDr at inference with Retrieval-Augmented Diagnosis (RAD): for a test image, similar cases are retrieved and summarized as contextual guidance in the prompt~\cite{he2024gsco}. 
\textbf{GSCo} (Generalist--Specialist Collaboration) is a two-stage framework that (i) builds a generalist GFM (MedDr) and lightweight specialist models, and (ii) performs collaborative inference via Mixture-of-Expert Diagnosis (MoED; using specialist predictions as references) and RAD (using specialists to retrieve similar cases)~\cite{he2024gsco}. GSCo evaluates across a large multi-dataset benchmark; we include GSCo as a strong InternVL-based baseline along with its MedDr and MedDr+RAD components~\cite{he2024gsco,chen2024internvl}.

\section{Additional Analysis}

\subsection{Analysis of Model Components}
\label{model_comp}
We analyze the contribution of each part in the model in Table \ref{tab:full_results_classification_analysis} and in Table \ref{tab:full_results_vqa_analysis}. The result of the first stage of training - the reader retrieval augmentation fine-tuning is the FT RAG then we check the contribution of training only the text retriever head and then finally is the contribution of training both the text and image retriever head as \methodname{}. We also explore the utility of converting the open question into a closed one in our retrieval loss. If we don't apply o3 model to convert the open question, we only trained on the closed one, which is referred to in Table \ref{tab:full_results_vqa_analysis} as Closed Question Only.  

\subsection{Performance with a State-of-the-Art LVLM Reranker}
\label{rerank_analysis_a}
We evaluate whether a state-of-the-art large vision–language model (LVLM), used as an optional reranker, can improve performance on inconsistent-retrieval predictions. Specifically, we test the o3 reasoning model \cite{o32025openai} as a reranker: given a query image to classify and four retrieved image–caption pairs, o3 is asked to select the single pair that is most informative for predicting the correct label. We compare this reranking to four alternatives: (i) using the top-1 retrieved candidate (no reranking), (ii) selecting the model’s prediction with the highest confidence (maximum logit), (iii) using o3 with captions only (no images), and (iv) the \methodname{} aggregation strategy.

Our results show that o3 generally outperforms the maximum-logit baseline but remains inferior to \methodname{}’s logit-fusion aggregation. We evaluated performance on the Breast, Derma, and Retina datasets. For the remaining datasets—VinDr-PCXR, BRSET, and the visual question answering (VQA) datasets—we did not run this evaluation because the images and/or retrieved candidates come from restricted-access sources (e.g., MIMIC-CXR, VinDr-PCXR, and BRSET). See results in Table~\ref{tab:rerankers}.

\begin{table*}[t!]
\centering

\begin{tabular}{ll cc cc cc cc cc}
\toprule
Backbone & Model & \multicolumn{2}{c}{Breast} & \multicolumn{2}{c}{Derma} & \multicolumn{2}{c}{Retina} & \multicolumn{2}{c}{VinDr} & \multicolumn{2}{c}{BRSET}  \\
\cmidrule(lr){3-4} \cmidrule(lr){5-6} \cmidrule(lr){7-8} \cmidrule(lr){9-10} \cmidrule(lr){11-12} 
& & ACC & F1 & ACC & F1 & ACC & F1 & ACC & F1 & ACC & F1  \\
\midrule
Qwen2-vl 
& \methodname{} N=2 & .87 & .83 & .74 & .50 & .63 & .50 & .56 & .14 & .47 & .35  \\
& \methodname{} N=4 & .87 & .84 & .76 & .50 & .65 & .50 & .57 & .14 & .49 & .37  \\
& \methodname{} N=6 & .87 & .84 & .75 & .50 & .63 & .52 & .58 & .14 & .48 & .37  \\
\midrule
Pixtral
& \methodname{} N=2 & .89 & .86 & .80 & .63 & .58 & .48 & .54 & .13 & .49 & .35  \\
& \methodname{} N=4 & .90 & .87 & .80 & .62 & .60 & .51 & .56 & .14 & .51 & .37  \\
& \methodname{} N=6 & .90 & .87 & .80 & .63 & .61 & .51 & .55 & .13 & .49 & .36  \\
\bottomrule
\end{tabular}

\caption{Ablation on the number of retrieved candidates (N) for classification tasks.}
\label{tab:ablation_classification}
\end{table*}

\begin{table*}[t!]
\centering

\begin{tabular}{ll cc cc cc }
\toprule
\textbf{Backbone} & Model & \multicolumn{2}{c}{VQA-RAD} & \multicolumn{2}{c}{SLAKE} & \multicolumn{2}{c}{PathVQA}  \\
\cmidrule(lr){3-4} \cmidrule(lr){5-6} \cmidrule(lr){7-8} 
& & Closed & Open & Closed & Open & Closed & Open  \\
\midrule
Qwen2-vl
& \methodname{} N=2 & .78 & .46 & .88 & .82 & .92 & .37  \\
& \methodname{} N=4 & .79 & .48 & .90 & .84 & .93 & .38  \\
& \methodname{} N=6 & .78 & .47 & .89 & .83 & .93 & .38  \\
\midrule
Pixtral
& \methodname{} N=2 & .78 & .45 & .89 & .83 & .93 & .38  \\
& \methodname{} N=4 & .78 & .47 & .90 & .84 & .93 & .37  \\
& \methodname{} N=6 & .78 & .47 & .89 & .84 & .93 & .38  \\
\bottomrule
\end{tabular}
\caption{Ablation on the number of retrieved candidates (N) for visual question answering tasks.}
\label{tab:ablation_vqa}
\end{table*}
\subsection{\methodname{} Variants with Medical Pre-Trained Backbones}
\label{medflamingo}

Although our work focuses on general-purpose LVLMs, we also evaluated a medical-specific baseline. We conducted the evaluation on four benchmarks—Breast, Derma, Retina, and VQA-RAD (closed-question subset). For the medical pre-trained baseline, we used Med-Flamingo and paired it with a medically pre-trained retriever (BiomedCLIP), using only one retrieval head (image retrieval without text retrieval). The performance gap between Med-Flamingo and our reader approach was small—0.01 in accuracy and 0.03 in F1. We hypothesize that this narrow gap may be due to Med-Flamingo's relatively older LLaMA backbone \cite{touvron2023llama}. Results are presented in Table~\ref{tab:appendix_full_results}. We also tested our general-purpose backbone with a medically pre-trained retriever, which showed improvements over FT RAG, with a similar pattern to the general-purpose retriever. Full results are presented in Table~\ref{tab:appendix_full_results_bio}.

\subsection{Different number of retrieved candidates}
\label{N_can}
Our model demonstrates robustness across varying numbers of retrieved candidates. We explore performance for N = 2, 4, and 6, showing consistent results across the different configurations. Results are presented in Table \ref{tab:ablation_classification} for classification tasks and in Table \ref{tab:ablation_vqa} for visual question answering tasks.

\section{Prompt Design}
\label{prompt_design}
We detail the prompts used in our experiments and analyses. To convert open questions into closed ones with the o3 model, we use the prompt in Prompt~\ref{lst:o3openprompt}. To assess the utility of o3 as a re-ranker of the most predictive image–caption pair (Section~\ref{rerank_analysis_a}, Appendix~D), we provide both LVLM backbones (Pixtral and Qwen2-VL) with the task-specific prompt in Prompt~\ref{lst:o3prompt}.

We also include the training prompts for \methodname{} across datasets—Prompts~\ref{lst:breast_our}, \ref{lst:derma_our}, \ref{lst:retina_our}, \ref{lst:BRSET}, \ref{lst:vindr_our}—and the VQA prompt (Prompt~\ref{lst:vqa_our}).

\lstset{
    basicstyle=\ttfamily\small,
    breaklines=true,
    frame=single,
    numberstyle=\tiny,
    captionpos=b,
    keepspaces=true,
    columns=flexible,
    breakatwhitespace=false,
    tabsize=3,
    float=H,         
    aboveskip=10pt,  
    belowskip=10pt,  
    abovecaptionskip=10pt,
    belowcaptionskip=10pt
}

\begin{figure*}[t!]
\begin{samepage}
\begin{lstlisting}[caption={Prompt used for o3 to convert open question into closed one}, label={lst:o3openprompt}]
    Convert the following open-ended question into a closed yes/no question based on the given answer.
The new question should be answerable with "{expected_answer}".

Original Question: {question}
Original Answer: {answer}
Expected New Answer: {expected_answer}

Please provide only the new closed question without any additional text or explanation.


\end{lstlisting}
\end{samepage}
\end{figure*}

\begin{figure*}[t!]
\begin{samepage}
\begin{lstlisting}[caption={Prompt used for retrieval candidate relevance evaluation}, label={lst:relevnatPrompt}]
    You are an expert medical AI assistant specializing in evaluating medical image and text relevance for clinical decision-making.

    Compare two medical (image, text) pairs and determine which is MORE RELEVANT to the ground truth label.
    
Ground Truth Label: "{gt_text}"

First Image: {image_1}
First Text: {text_1}

Second Image: {image_2}
Second Text: {text_2}

Consider:
- Medical image findings
- Medical terminology accuracy in text
- Clinical relevance to the ground truth
- Diagnostic information alignment
- How useful each (image, text) pair would be for the given condition

Return JSON format:
{{
    "choice": <1 or 2>,
    "confidence": <"high", "medium", or "low">,
    "explanation": "<brief 1-sentence explanation>"
}}

Be concise and accurate. Choose the option that is MORE medically relevant to the ground truth.

\end{lstlisting}
\end{samepage}
\end{figure*}

\begin{figure*}[t!]
\begin{samepage}
\begin{lstlisting}[caption={Prompt used for o3}, label={lst:o3prompt}]
    Task:
    
    You are an expert assistant selecting the most informative reference for medical image classification.
    
    Description:
    A patient's [MRI/CT/X-ray] image needs to be classified into one of the following categories: [List of possible labels].
    You are given four candidate reference pairs, each consisting of a caption and its associated image, drawn from PubMed literature.
    Your goal is to determine which single candidate provides the most useful information to help an AI model correctly classify the provided image.
    
    Input Information:
    - Medical image: The patient's MRI/CT/X-ray image is provided here.
    - Classification task: Classify the image into one of the following categories: [List of possible labels].
    - Candidates:
        - Candidate 0:
            - Caption: [Caption 0]
            - Image: [Image 0]
        - Candidate 1:
            - Caption: [Caption 1]
            - Image: [Image 1]
        - Candidate 2:
            - Caption: [Caption 2]
            - Image: [Image 2]
        - Candidate 3:
            - Caption: [Caption 3]
            - Image: [Image 3]
    
    Instructions:
    1. For each candidate, carefully assess how informative its caption and image are for the current classification task.
    2. Specifically, evaluate whether the candidate describes imaging features, findings, or clinical context relevant to distinguishing among the listed categories.
    3. Compare all candidates and select the one that gives the clearest, most discriminative information to aid the classification.
    4. Output only the number of the selected candidate in the following format:
       Caption Number: [your selected number]


\end{lstlisting}
\end{samepage}
\end{figure*}

\begin{figure}[t!]
\begin{samepage}
\begin{lstlisting}[caption={Prompt used by \textsc{Jomed} for Breast classification}, label={lst:breast_our}]
        <retrieved image>background:<retrieved text>\n\n<query image>Does this breast ultrasound image show signs of cancer?
\end{lstlisting}
\end{samepage}
\end{figure}

\begin{figure}[t!]
\begin{samepage}
\begin{lstlisting}[caption={Prompt used by \textsc{Jomed} for Retina classification}, label={lst:retina_our}]
        <retrieved image>background:<retrieved text>\n<query image>what is the severity of diabetic retinopath?
\end{lstlisting}
\end{samepage}
\end{figure}
\begin{figure*}[t!]
\begin{samepage}
\begin{lstlisting}[caption={Prompt used by \textsc{Jomed} for Derma classification}, label={lst:derma_our}]
        <retrieved image>background:<retrieved text>\n\nProvide answer according to the labels:\n"
          "0 - 'Actinic keratoses and intraepithelial carcinoma'\n"
          "1 - 'Basal cell carcinoma'\n"
          "2 - 'Benign keratosis-like lesions'\n"
          "3 - 'Dermatofibroma'\n"
          "4 - 'Melanoma'\n"
          "5 - 'Melanocytic nevi'\n"
          "6 - 'Vascular lesions'\n\n"<query image>What type of skin lesion does the patient have?
\end{lstlisting}
\end{samepage}
\end{figure*}


\begin{figure*}[t!]
\begin{samepage}
\begin{lstlisting}[caption={Prompt used by \textsc{Jomed} for Vindr-PCXR classification}, label={lst:vindr_our}]
       <retrieved image>background:<retrieved text> Provide answer according to the labels:\n
       0 - 'No finding'\n
       1 - 'Bronchitis'\n
       2 - 'Brocho-pneumonia'\n
       3 - 'Other disease'\n
       4 - 'Bronchiolitis'\n
       5 - 'Situs inversus'\n
       6 - 'Pneumonia'\n
       7 - 'Pleuro-pneumonia'\n
       8 - 'Diagphramatic hernia'\n
       9 - 'Tuberculosis'\n
       10 - 'Congenital emphysema'\n
       11 - 'CPAM'\n
       12 - 'Hyaline membrane disease'\n
       13 - 'Mediastinal tumor'\n
       14 - 'Lung tumor'\n\n
       \n\n<image>Question: Look at this X-ray scan and select all abnormalities you see from the given labels Answer:
\end{lstlisting}
\end{samepage}
\end{figure*}

\begin{figure*}[t!]
\begin{samepage}
\begin{lstlisting}[caption={Prompt used by \textsc{Jomed} for BRSET classification}, label={lst:BRSET}]
<retrieved image>background:<retrieved text> Provide answer according to the labels:\n
0 - 'no findings'\n
1 - 'diabetic_retinopathy'\n
2 - 'macular_edema'\n
3 - 'scar'\n
4 - 'nevus'\n
5 - 'amd'\n
6 - 'vascular_occlusion'\n
7 - 'hypertensive_retinopathy'\n
8 - 'drusens'\n
9 - 'hemorrhage'\n
10 - 'retinal_detachment'\n
11 - 'myopic_fundus'\n
12 - 'increased_cup_disc'\n
13 - 'other'\n\n
\n\n<image>Question: Look at retinal fundus image and select all abnormalities you see from the given labels 
\end{lstlisting}
\end{samepage}
\end{figure*}

\begin{figure}[t!]
\begin{samepage}
\begin{lstlisting}[caption={Prompt used by \textsc{Jomed} for visual question answering tasks}, label={lst:vqa_our}]
        <retrieved image>background:<retrieved text>\n<query image><query question>?
\end{lstlisting}
\end{samepage}
\end{figure}


\end{document}